\documentclass{article}

\usepackage{arxiv}

\usepackage[utf8]{inputenc} % allow utf-8 input
\DeclareUnicodeCharacter{2212}{-}
\usepackage[T1]{fontenc}    % use 8-bit T1 fonts
\usepackage{hyperref}       % hyperlinks
\usepackage{url}            % simple URL typesetting
\usepackage{booktabs}       % professional-quality tables
\usepackage{amsfonts}       % blackboard math symbols
\usepackage{amsmath,amsfonts,amssymb,amsthm}
\usepackage{nicefrac}       % compact symbols for 1/2, etc.
\usepackage{microtype}      % microtypography
\usepackage{lipsum}
\usepackage{multirow,tabu,array,booktabs}
\usepackage{graphicx}
\usepackage[table,xcdraw]{xcolor}
\usepackage{multicol}
\usepackage{float}

\def\correction#1{%
	\abovedisplayshortskip=#1\baselineskip\relax\belowdisplayshortskip=#1\baselineskip\relax%
	\abovedisplayskip=#1\baselineskip\relax\belowdisplayskip=#1\baselineskip\relax}

\fboxsep=\tabcolsep\relax
\fboxrule=\arrayrulewidth\relax 

\title{An empirical comparison between stochastic and deterministic centroid initialisation for K-Means variations}

\author{
  Avgoustinos Vouros%\thanks{Use footnote for providing further information about author (webpage, alternative address)---\emph{not} for acknowledging funding agencies.} 
  \\
  Department of Computer Science\\
  University of Sheffield\\
  Sheffield S10 2TN, UK \\
  \texttt{avouros1@sheffield.ac.uk} \\
   \And
  Stephen Langdell \\
  Numerical Algorithms Group (NAG) \\
  Wilkinson House, Jordan Hill Road,\\ Oxford OX2 8DR, UK \\
   \And
  Mike Croucher \\
  Numerical Algorithms Group (NAG) \\
  Wilkinson House, Jordan Hill Road,\\ Oxford OX2 8DR, UK \\
	\And     
  Eleni Vasilaki \\
  Department of Computer Science\\
  University of Sheffield\\
  Sheffield S10 2TN, UK \\
}

\begin{document}
\maketitle

\begin{abstract}
		K-Means is one of the most used algorithms for data clustering and the usual clustering method for benchmarking. Despite its wide application it is well-known that it suffers from a series of disadvantages; it is only able to find local minima and the positions of the initial clustering centres (centroids) can greatly affect the clustering solution. Over the years many K-Means variations and initialisation techniques have been proposed with different degrees of complexity. In this study we focus on common K-Means variations along with a range of deterministic and stochastic initialisation techniques. We show that, on average, more sophisticated initialisation techniques alleviate the need for complex clustering methods. Furthermore, deterministic methods perform better than stochastic methods. However, there is a trade-off: less sophisticated stochastic methods, executed multiple times, can result in better clustering. Factoring in execution time, deterministic methods can be competitive and result in a good clustering solution. These conclusions are obtained through extensive benchmarking using a range of synthetic model generators and real-world data sets.
\end{abstract}

% keywords can be removed

	\section{Introduction}

The most well-known algorithm in the field of clustering analysis is the K-Means algorithm. Its simplicity, versatility and efficiency makes it popular in many different research fields \cite{pena1999empirical,jain2010data}. Despite its reputation and success in many different studies, it has a series of disadvantages such that it can detect only spherical and well-separated clusters, it is sensitive to outliers, highly dependent on the features (dimensions) of the data set and it only converges to local minima \cite{jain2010data}. Over the years a number of K-Means variations (Lloyd's K-Means \cite{slonim2013hartigan}, Hartigan-Wong's K-Means \cite{hartigan1975clustering}), K-Means inspired algorithms (K-Medians \cite{charu2013data}), and K-Means initialisation methods \cite{celebi2013comparative} have been proposed in order to overcome some of these issues. Such methods have also enhanced KMeans with additional properties such as feature selection mechanisms \cite{witten2010framework,kondo2016rskc}, semi-supervised capabilities \cite{bilenko2004integrating,vouros2021semi}, and outliers robustification \cite{al2009robust,brodinova2017robust}. 

In the literature there are various studies regarding the importance of the initial selection of cluster centroids for the performance of the K-Means algorithm \cite{jain2010data} and extensive testing on various initialisation techniques \cite{celebi2013comparative,franti2019much}, but a detailed comparison on the effects on these techniques on common K-Means variations is not available. We hypothesize that sophisticated initialisation methods alleviate the need for complex clustering and, if deterministic, they could lead to satisfactory solutions within a single execution of the clustering algorithm. Consequently, they would alleviate the need for executing a stochastic method multiple times and picking the best clustering based on some criterion. 

In order to investigate this hypothesis we compare different clustering initialisation methods, namely Random \cite{macqueen1967some}, K-Means++ \cite{arthur2007k}, Maximin \cite{gonzalez1985clustering} ROBust INitialisation (ROBIN) \cite{al2009robust}, Kaufman \cite{kaufman2009finding} and Density K-Means++ (DK-Means++) \cite{nidheesh2017enhanced}, and their effects on common K-Means variations, Lloyd's K-Means \cite{jain2010data}, Hartigan-Wong's K-Means \cite{hartigan1979algorithm,hartigan1975clustering} and K-Medians \cite{charu2013data}. We show that more sophisticated initialisation methods reduce on average the performance difference among the K-Means implementations and that the deterministic DK-Means++ method can achieve better average performance than stochastic methods. Nevertheless there is a trade-off, simplistic stochastic methods can achieve better clustering performance if executed multiple times due to the potential of discovering better local minima. For very large data sets where execution time is a factor, a single run using a deterministic initialisation method can be competitive compared to multiple runs using stochastic initialisation methods. 
%In more detail, we find that deterministic versions (Maximin(D) \cite{katsavounidis1994new} and ROBIN(D) \cite{al2009robust}) of stochastic algorithms (Maximin(S)  \cite{gonzalez1985clustering} and ROBIN(S) \cite{brodinova2017robust}) are having equal or better performance than the average performance of their stochastic versions but worst when the latter are executed multiple times. 
%However, we also show that by using an unsupervised criterion to select the best performing clustering of a stochastic algorithm can lead to a better performance than the single run of deterministic algorithm, due to the potential of discovering better local minima. 
A similar study comparing many different intialisation methods has been performed by \cite{celebi2013comparative} but it is focused on algorithms of linear complexity without considering various K-Means implementations. Recently, another study \cite{franti2019much} was performed on stochastic initialization heuristics for K-Means and on how much the algorithm can be improved by repetition. They based their conclusions on a clustering benchmark \cite{franti2018k,franti2019much} which contains standalone data sets with different properties and they showed that K-Means performance is in general poor on unbalanced data sets and that the algorithm is not affected by high dimensionality while more iterations can improve its performance on overlapping clusters. In our case we performed a more extensive benchmarking by taking into consideration data set generation models as well as standalone data sets. The models gave us the ability to perform hypothesis testing in order to strengthen our conclusions and to account for variability.

The code of the clustering methods, data set model generators, scripts and a standalone application to reproduce this research are available in the GitHub repository \url{https://github.com/avouros/Code-KMeans-benchmark} (under the branch additions).

\section{Material and method}

\subsection{The K-Means algorithm}
Given K initial centroids, the K-Means algorithm \cite{jain2010data} assigns the data points into $K$ clusters in a way that minimizes the within cluster sum of squares (WCSS):
\begin{align}\label{kmeans}
	WCSS &= \sum_{k=1}^{K}\sum_{\binom{i=1}{x_{i:} \in c_k}}^{n_k} \sum_{j=1}^{p}(x_{ij}-m_{kj})^2,
\end{align}	
where $K$ is the number of clusters, $n_k$ the number of data points (observations) of the $k$-th cluster and $p$ the dimensionality (number of features) of a given dataset; $x_{ij}$ is the value of the $j$-th feature of the $i$-th data point, $x_{i:}$ is the vector representing the $i$-th datapoint; $m_{k:}$ specifies the location of the $k\text{-}th$ cluster centroid. This problem is equivalent to maximizing the between cluster sum of squares (BCSS) which is given by \cite{witten2010framework}:
\begin{align}\label{kmeans3}
	BCSS &= \sum_{j=1}^{p} \bigg( \sum_{i=1}^{n}(x_{ij}-\mu_{1j})^2 - \sum_{k=1}^{K}\sum_{\binom{i=1}{x_{i:} \in c_k}}^{n_k}(x_{ij}-m_{kj})^2 \bigg),	
\end{align}	
where $\mu_{1:}$ specifies the global centroid assuming that all the data points belong to one cluster.

\subsubsection{Lloyd's K-Means}
Lloyd's method is the most commonly used K-Means and the standard K-Means clustering method in many programming languages such as MATLAB \cite{MATLAB2019} and Python \cite{Python}. The steps of this algorithm are as follows \cite{jain2010data}:
\begin{enumerate}
	\item Initialise $K$ initial centroids $m_{1j}, \dots, m_{Kj}$ using some initialisation method.
	
	\item Assign each data point $x_{i:}$ to cluster $c_{k^*}$,
	\begin{align}\label{lkmeansobj}
		k^* &= \underset{k}{argmin}\bigg\{ \sum_{j=1}^{p} (x_{ij}-m_{kj})^2    \bigg\}.
	\end{align}	
	
	\item Recompute each cluster centroid using the formula,
	\begin{align}
		m_{kj} = \frac{\sum_{\binom{i=1}{x_{i:} \in c_k}}^{n_k} x_{ij}}{n_k}.
	\end{align}	
	
	\item Go to step 2 until converge.
\end{enumerate}

\subsubsection{Hartigan-Wong's K-Means}
Hartigan-Wong's K-Means algorithm is an alternative to Lloyd's K-Means and the default K-Means of the R language \cite{Rlanguage}. In the study of \cite{slonim2013hartigan} it is shown that this method has lower probability of converging to a local minima solution compared to Lloyd's method in exchange of extra complexity. The steps of the algorithm are as follows \cite{hartigan1975clustering,slonim2013hartigan}:
\begin{enumerate}
	\item Initialise $K$ initial centroids $m_{1j}, \dots, m_{Kj}$ using an initialisation method.
	
	\item Assign each data point $x_{i:}$ to cluster $c_{k'}$,
	\begin{align}\nonumber
		k' &= \underset{k}{argmin}\bigg\{  \sum_{j=1}^{p} (x_{ij}-m_{kj})^2    \bigg\}.
	\end{align}		
	
	\item Set an indicator $s=1$.			
	
	\item For each data point $x_{i:}$
	
	\begin{enumerate}
		\item Remove it from its cluster $c_{k'}$.
		\item Compute the centroid of $c_{k'}$ using the remaining points in that cluster,
		\begin{align}\nonumber
			m_{k' j} = \frac{1}{n_{k'}}\sum_{\binom{i=1}{x_{i:} \in c_{k'}}}^{n_{k'}} x_{ij}.
		\end{align}	
		\item Assign $x_{i:}$ to cluster $c_{k^*}$,
		\begin{align}\nonumber
			k^* &= \underset{k}{argmin}\bigg\{ \sum_{j=1}^{p} (x_{ij}-m_{kj})^2    \bigg\}.
		\end{align}					
		\item Recompute the centroid of the cluster $c_{k^{*}}$,
		\begin{align}\nonumber
			m_{k^{*}j} = \frac{1}{n_{k^{*}}}\sum_{\binom{i=1}{x_{i:} \in c_{k^*}}}^{n_{k^{*}}} x_{ij}.
		\end{align}
		\item If $k^* \neq k'$ set $s=0$.	
	\end{enumerate}	
	
	\item If $s=0$, set $s=1$ and go to step 4.
	
\end{enumerate}	

\subsubsection{The K-Medians algorithm}
The K-Medians algorithm \cite{charu2013data} is similar to the K-Means but uses the median instead of the mean to calculate the cluster centroid. The objective function of the algorithm is given by the equation:
\begin{align}\label{kmed1}
	E &= \sum_{k=1}^{K}\sum_{\binom{i=1}{x_{i:} \in c_k}}^{n_k} \sum_{j=1}^{p} |x_{ij}-\bar{m}_{kj}|,
\end{align}	
where $\bar{m}_{kj}$ specifies the location of the $k\text{-}th$ cluster centroid in the $j$-th dimension which is computed by taking the median of the data points $x_{ij}$ belonging to that cluster. K-Medians corresponds to the $L_1$-norm as opposed to the $L_2$-norm of K-Means \cite{charu2013data}. The use of median in place of the mean makes the K-Medians algorithm robust to outliers \cite{feldman2012data,whelan2015understanding} since the median has a breaking point of 0.5, i.e. even if half of the data set is corrupted by outliers the median of the corrupted data set will be similar to the median of the original data set \cite{lopuhaa1991breakdown}. The common implementation of the algorithm is similar to Lloyd's K-Means where, in the $3rd$ step of Lloyd's algorithm, the median is used to compute the new centroids locations instead of the mean., i.e. $\forall \hphantom{x} k$ $\bar{m}_{k:} = median(\{x_{i:}\}), \hphantom{x} x_{i:} \in c_k$. %It should be noted that during the $2nd$ step of the algorithm (the assignment step) the use of Euclidean or Euclidean$^2$ will have qualitatively the same result.

\subsection{K-Means initialisation methods}
Let $D(x_{i:})$ denote the distance between data point $x_{i:}$ and the nearest of the selected cluster centroids, $m_{k:}$, $k = 1, \dots, L$, with $L$ being the number of selected centroids ($L\leq K$): 
\begin{align}\label{dxi}
	D(x_{i:}) = \underset{k}{min} \sqrt{\sum_{j=1}^{p} (x_{ij}-m_{kj})^2}. 
\end{align}

\subsubsection{Random}
The initialisation method of MacQueen \cite{macqueen1967some} proposes a random selection of data points from the data set which will be the initial centroids. This is one of the earliest clustering initialisation techniques and an improvement of Jancey's method \cite{jancey1966multidimensional}. The latter study suggested the centroids to be at random locations within the minimum hypersphere of the data set but this might result in empty clusters to be generated after the execution of the K-Means algorithm.

\subsubsection{K-Means++}
K-Means++ \cite{arthur2007k} is a standard clustering initialisation technique in many programming languages such as MATLAB and Python. It has linear complexity $\mathcal{O}(N)$ and it uses a probabilistic approach in order to select as initial centroids data points that are far away from each other. The steps of this algorithm are as follows:
\begin{enumerate}
	\item Select randomly a data point $x_{i:}$ as the first centroid $m_{1:}$ and set $k=2$.
	\item Choose another data point $x_{i':}$ as the next centroid $m_{k:}$ with probability 
	\begin{align}\nonumber
		p(x_{i':})=\frac{D(x_{i':})^2}{\sum_{i=1}^{n}D(x_{i:})^2}
	\end{align}
	and set $k=k+1$.
	\item While $k \leq K$ go to step 2.
\end{enumerate}

\subsubsection{Maximin}
The maximin method of \cite{gonzalez1985clustering} picks data points as cluster centroids that are far apart form each other.
\begin{enumerate}
	\item Select randomly a data point $x_{i:}$ as the first centroid $m_{1:}$ and set $k=2$.
	\item Select as the next centroid $m_{k:}=x_{i':}$ with $i'= \underset{i}{argmax}\{D(x_{i:})\}$ and set $k=k+1$.
	\item While $k \leq K$ go to step 2.
\end{enumerate}	
Maximin has linear complexity $\mathcal{O}(N)$. The study of \cite{katsavounidis1994new} proposed a modification in the first step of the algorithm to select as the first centroid the data point with the maximum Euclidean norm \cite{celebi2013comparative}. In this way the method can become deterministic.

\subsubsection{Kaufman}
Kaufman and Rousseeuw \cite{kaufman2009finding} proposed a deterministic method for centroids initialisation. Their method is as follows \cite{pena1999empirical}:
\begin{enumerate}
	\item Select the closest data point to the global centroid of the data set as the first centroid $m_{1:}$ and set $k=2$.
	\item For every two non-selected data points $x_{i:}$ and $x_{i':}$ calculate,
	\begin{align}\nonumber
		C_{i'i} = max \Bigg\{D(x_{i:}) - \sqrt{\sum_{j=1}^{p} (x_{ij}-x_{i'j})^2} , 0 \Bigg\}.
	\end{align}
	\item Select as the next centroid  $m_{k:}=x_{i^*:}$, with $i^*= \underset{i}{argmax}\{\sum_{i'}C_{i'i}\}$ and set $k=k+1$.
	\item While $k \leq K$ go to step 2.
\end{enumerate}
Kaufman's and Rousseeuw's algorithm has quadratic complexity $\mathcal{O}(N^2)$ because of the computation of the pairwise distances \cite{celebi2013comparative}. 

\subsubsection{ROBust INitialisation (ROBIN)}\label{ssROBIN}
ROBIN \cite{al2009robust} is a robust to outliers initialisation method. It uses the Local Outlier Factor (LOF) \cite{breunig2000lof} in order to select as initial centroids data points that are far away from each other and also representative points of dense regions in the data set. In addition it requires one more tuning parameter which is the number of neighboring data points $mp$ to be consider when computing the LOF of each data point. The LOF score of each data point, $LOF(x_{i:},mp)$, is given by the algorithm below \cite{al2009robust} (where $N(x_{i:},mp)$ is the set of the $mp$ nearest data points to the $x_{i,:}$ data point, with $|N(x_{i:},mp)| \geq mp$):

\begin{enumerate}
	\item Compute the density of each data point $x_{i:}$,
	\begin{equation}
		density(x_{i:},mp) = \frac{|N(x_{i:},mp)|}{ \sum_{x_{i':} \in N(x_{i:},mp)}\sqrt{\sum_{j=1}^{p}(x_{ij}-x_{i'j})^2} } , i \neq i' \label{form1}.
	\end{equation} 	
	\item Compute the average relative density of each data point $x_{i:}$,
	\begin{equation}
		ard(x_{i:},mp) = \frac{density(x_{i:},mp)}{\frac{\sum_{x_{i':} \in N(x_{i:},mp)}density(x_{i':},mp)}{|N(x_{i:},mp)|}} .\label{form2}
	\end{equation} 	
	\item Compute the LOF score of each data point $x_{i:}$,
	\begin{equation}
		LOF(x_{i:},mp) = \frac{1}{ard(x_{i:},mp)}.\label{form3}
	\end{equation} 	
\end{enumerate}
The ROBIN algorithm ($ K>1$ ) is described below \cite{al2009robust}:
\begin{enumerate}
	\item Pick a reference data point $x_{r:}$ from the data set.
	\item Sort the data points in decreasing order of their distance from $x_{r:}$.
	\item For each $x_{i:}$ in sorted order, pick the first data point $x_{i':}$ for which \\$LOF(x_{i':},mp) \approx 1$ as the first centroid $m_{1:}$ and set $k=2$.
	\item Sort the data points in decreasing order based on $D(x_{i:})$. For the formula of $D(x_{i:})$ refer to equation \ref{dxi}.
	\item For each $x_{i:}$ in sorted order, pick the first data point $x_{i':}$ for which \\$LOF(x_{i':},mp) \approx 1$ as the next centroid $m_{k:}$ and set $k=k+1$.
	\item While $k \leq K$ go to step 4.
\end{enumerate}

The computational cost of this method is dominated by the complexity of sorting, which is $\mathcal{O}(N \textit{ log } N)$ \cite{celebi2013comparative} but for the LOF score calculation we have a choice of algorithms varying from $\mathcal{O}(N)$ to $\mathcal{O}(N^2)$, that can be chosen based on dimensionality-related constraints, see \cite{breunig2000lof}. Regarding the $4^{th}$ step of the algorithm, in an R implementation (refer to the study of \cite{brodinova2017robust}) the formula $LOF(x_{r':},mp) < 1.05$ was used but since the LOF score can also be lower than $1$, in our experiments, we used the formula $1 - e < LOF(x_{r':},mp) < 1 + e$ where $e$ was set to $0.05$.	In the original algorithm \cite{al2009robust} the authors are using the algorithm in a deterministic manner by setting the reference point on step 2, $x_r:$ to the origin. In the $R$ implementation of \cite{brodinova2017robust} the reference point is chosen at random. In this study we test both methods, \textit{ROBIN(S)} will refer to the stochastic method of \cite{brodinova2017robust} while \textit{ROBIN(D)} will refer to the deterministic method of \cite{al2009robust}.

\subsubsection{Density K-Means++ (DK-Means++)}
DK-Means++ \cite{nidheesh2017enhanced} is a deterministic method for centroids initialisation based on data density. It is an improved method of \cite{rodriguez2014clustering,lan2015density} since it requires only to define the number of clusters $K$ and utilizes a heuristic to detect dense regions in the data set based on a radius $\varepsilon$ in order to select optimal centroids. The radius $\varepsilon$ can be computed form the following algorithm \cite{nidheesh2017enhanced}:
\begin{enumerate}
	\item Construct the minimum spanning tree of the distance matrix of the data set.
	\item Let $\Lambda$ be the weights of edges of the Minimum Spanning Tree and IQR the Inter Quartile Range. Then,
	\begin{align}\nonumber
		\varepsilon = 3\cdot IQR(\Lambda) + 75^{th} percentile(\Lambda).
	\end{align}
\end{enumerate}
The DK-Means++ algorithm is described below \cite{nidheesh2017enhanced}:
\begin{enumerate}
	\item Compute the the local density $p(x_{i:})$ of each data point using the formula: 
	\begin{align}\nonumber
		p(x_{i:}) = \sum_{x_{i'} \in \varepsilon\text{-}neighbors(x_{i:})}\exp\Bigg(\frac{- \sqrt{\sum_{j=1}^{p}(x_{ij}-x_{i'j})^2 }}{\varepsilon}\Bigg).		
	\end{align}
	where $\varepsilon\text{-}neighbors(x_{i:})$ are the data points falling under the hypersphere with centroid $x_{i:}$ and radius $\varepsilon$.
	\item Normalize $p(x_{i:})$ using the \textit{min-max} normalization.
	\item The first cluster centroid $m_{1:}$ is the data point $x_{i^*:}$ for which $p(x_{i^*:}) = max\{p(x)\}$. Then $m_{1:}=x_{i^*:}$  and  $k = 2$.
	\item Compute the prospectiveness all data points that are not selected as centers given the formula, $\phi(x_{i:}) = p(x_{i:})\cdot D(x_{i:})$. For the formula of $D(x_{i:})$ refer to equation \ref{dxi}.
	\item The next centroid $m_{k:}$ is the data point with maximum prospectiveness:  $m_{k:}=x_{i^*:}$ with $i^*=\underset{i}{argmax}\{\phi(x_{i:})\}$ and $k=k+1$.
	\item While $k \leq K$ go to step 4.
\end{enumerate}
The computation of $\varepsilon\text{-}neighbors$ contributes to the complexity of DK-Means++. It is dominated by the distance matrix computation, which is $O(n^2)$. The computation of the Minimum Spanning Tree depends of the algorithm used to compute it, efficient implementations of Kruskal's algorithm and Prim's algorithm require $O$($n$ log $n$) time to compute a minimum spanning tree \cite{moret1992empirical}. As $O(n^2)$ is the dominating complexity, the entire process of finding initial centroids, of DK-Means++ has a time complexity of $O(n^2)$.
%to $\mathcal{O}\big((2n-1) \textit{ log } n\big)$ depending on the algorithm used to compute the minimum spanning tree of the distance matrix (e.g. Kruskal's or Prim's algorithm) \cite{moret1992empirical}. Since the distance matrix is a fully connected graph it will have vertices $V = n$ and edges $E = n-1$, where $n$ is the number of data points in the data set. In case of Kruskal's algorithm, which has complexity $\mathcal{O}\big((n-1) \textit{ log } n\big)$, the total algorithm complexity will be dominated by the sorting required in step 5 of the algorithm thus $\mathcal{O}(n \textit{ log } n)$ \cite{celebi2013comparative}. In case of Prim's algorithm the complexity can reach $\mathcal{O}\big((2n-1) \textit{ log } n\big)$ depending on its implementation. If the distance matrix is unknown then the algorithm complexity is dominated by the distance matrix computation which is $\mathcal{O}\big(\frac{1}{2} n(n-1)\big)$.

\subsection{Clustering evaluation}

In the literature there are many indexes for assessing the clustering performance \cite{al2009robust,rendon2011internal,franti2018k}. Here, in order to evaluate the clustering solutions, we selected to use one external (supervised) criterion, the Purity and one internal (unsupervised) criterion, the Silhouette index. 

\subsubsection{Purity}

Let our data belong to different classes $\ell_\mathsf{k}$ and that the number of classes $\mathsf{K}$ equals the number of clusters $K$, for each cluster the purity is defined as \cite{rendon2011internal}, 
\begin{equation}\label{cp1}
	P_{c_k} = \frac{1}{n_{k}}\underset{\mathsf{k}}{max}\{ n_{\ell_\mathsf{k}}^{(k)} \},
\end{equation}
where $\underset{\mathsf{k}}{max}\{ n_{\ell_\mathsf{k}}^{(k)} \}$ specifies the number of data points of the dominant class of the $k$-th cluster. The overall clustering purity index is then computed as,
\begin{equation}\label{cp2}
	P = \sum_{i=1}^{K}\frac{n_{k}}{n}P_{c_k}.
\end{equation}
The purity index is bounded between $(0 \text{ } 1]$; larger values of purity correspond to better performance accuracy and a purity of $1$ specifies an accuracy of $100\%$ meaning that each cluster has data points from only one class.

%	\subsubsection{Adjusted rand index (ARI)}
%	
%	The adjusted rand index \cite{yeung2001details} is a measurement of similarity between the ground truth of the data set and a partition given from an algorithm. Considering pairs of data points then the total pairs in a data set of $n$ elements is $\begin{pmatrix}n\\2\end{pmatrix}$. Let $n_{lk}$ be common elements in the ground truth cluster $l$ and the cluster $k$, $n_{l}$ the number of elements in the ground truth cluster $l$ and  $n_{k}$ the number of elements in the cluster $k$. Then the adjusted rand index is defined by equation \ref{eqari},
%	\begin{equation}\label{eqari}
%		\frac {\sum_{lk}\binom{n_{lk}}{2} - \big{[}\sum_{l}\binom{n_{l}}{2} \sum_{k}\binom{n_{k}}{2}\big{]} / \binom{n}{2}} {\frac{1}{2} \big{[}\sum_{l}\binom{n_{l}}{2} + \sum_{k}\binom{n_{k}}{2}\big{]} - \big{[}\sum_{l}\binom{n_{l}}{2} \sum_{k}\binom{n_{k}}{2}\big{]} / \binom{n}{2}}
%	\end{equation}
%	This particular index has also been used in previous studies \cite{yeung2001empirical,al2009robust,brusco2017comparison} for benchmarking purposes. 

\subsubsection{Silhouette index}

The silhouette index is a value that specifies the degree of similarity between a data point and other data points of the same cluster and the dissimilarity between a data point and other data points in different clusters. The silhouette index of the data point $x_{i:} \in c_{k}$ is given by \ref{silh1} \cite{rousseeuw1987silhouettes},
\begin{equation}\label{silh1}
	S_{x_{i:}} = \frac{b_{x_{i:}}-a_{x_{i:}}}{max\{b_{x_{i:}},a_{x_{i:}}\}},
\end{equation}	
where $a_{x_{i:}}$ is the average distance of $x_{i:}$ and all the other data points in the cluster that $x_{i:}$ belongs to
\begin{equation}\label{silh1a}
	a_{x_{i:}} = \frac{1}{n_k-1} \sum_{\binom{i'=1}{x_{i':} \in c_k}}^{n_k} \sqrt{\sum_{j=1}^{p}(x_{ij}-x_{i'j})^2}
\end{equation}	
and $b_{x_{i:}}$ is the minimum average distance of $x_{i:}$ to all the other data points in other clusters,
\begin{equation}\label{silh1b}
	b_{x_{i:}} = \underset{k'}{min} {\frac{1}{n_{k'}} \sum_{\binom{i'=1}{x_{i':} \in c_{k'}}}^{n_{k'}} \sqrt{\sum_{j=1}^{p}(x_{ij}-x_{i'j})^2}, k' \neq k}.
\end{equation}	
We can then define the average silhouette index,
\begin{equation}\label{silh3b}
	S = \frac{1}{n}\sum_{i=1}^{n}S_{x_{i:}}.
\end{equation}
The silhouette index is commonly used to estimate the number of clusters in a data set but it can also be used to assess the clustering quality \cite{rousseeuw1987silhouettes}. The Silhouette index is bounded between $-1$ and $1$, where $1$ specifies maximum separation of clusters and maximum within cluster density while other indexes, such as the distortion score, gives only an estimation of the latter.

\section{Benchmarks}

\begin{figure}
	\centering
	\includegraphics[width=\linewidth]{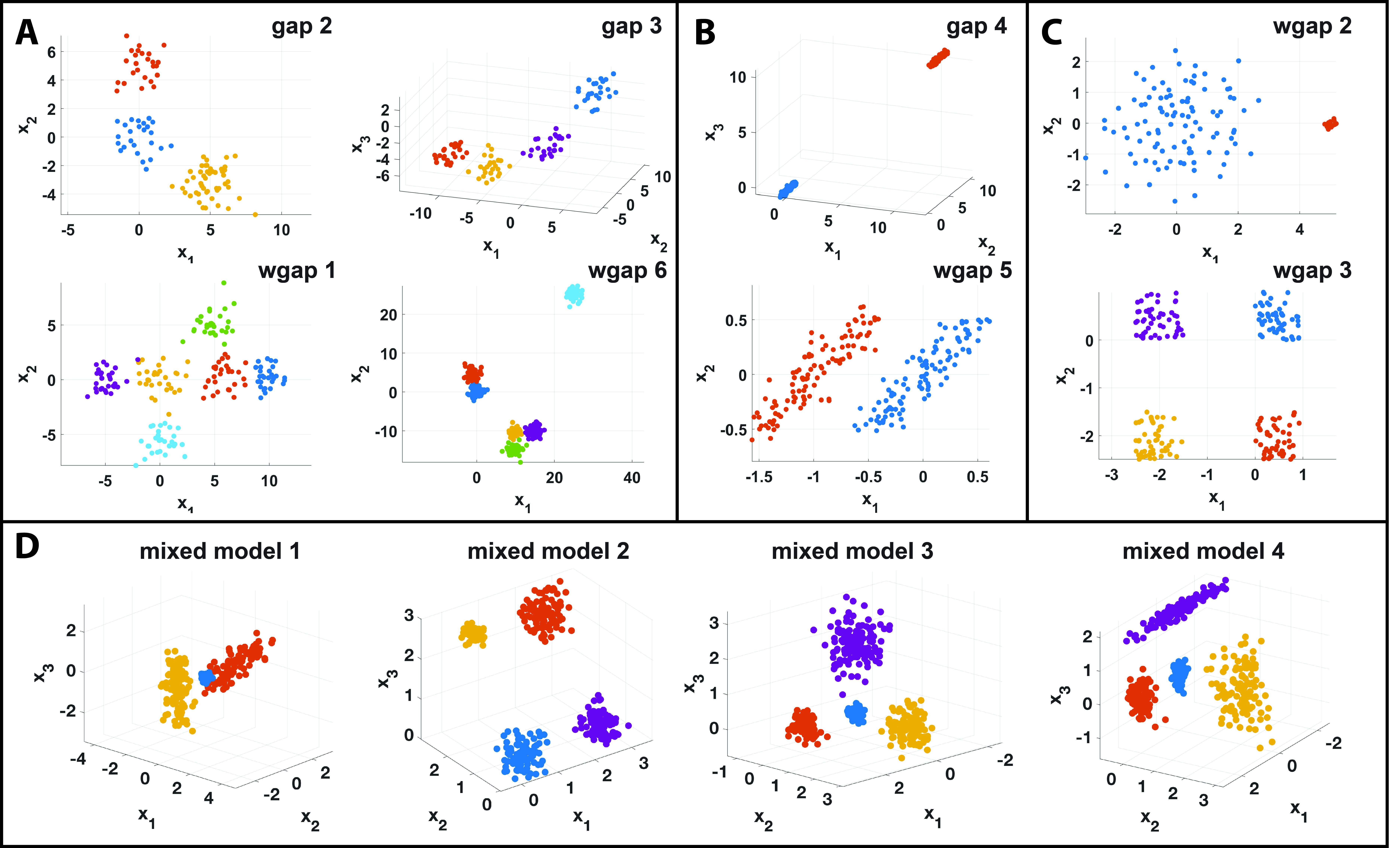}
	\caption{\textbf{Model visualization.} Examples of data used in this study. Gap (gap) and weighted gap (wgap) models are separated into three categories: \textbf{(A)} 4 Gaussian models, \textbf{(B)} 2 elongated models and \textbf{(C)} one highly unbalanced model (wgap2) and one non-Gaussian model (wgap3) in which the clusters are generated from the exponential distribution. \textbf{(D)} Mixed models: these are the additional models proposed in our study that contain clusters with mixed properties such as different sizes (unbalanced) and/or generated from Gaussian and non-Gaussian distributions.} \label{variousModels}  
\end{figure}

In our experiments we use the synthetic data from the studies of Tibshirani et al. (gap statistic) \cite{tibshirani2001estimating}, Yan and Ye (weighted gap statistic) \cite{yan2007determining} and Brodinova et al. \cite{brodinova2017robust}. A summary of the models can be found in Table \ref{gdatasets} grouped by specific properties of the models. For more information refer to the relevant studies and also to Figure \ref{variousModels} for a sample visualization of each model. We exclude model 1 from the gap statistic study \cite{tibshirani2001estimating} since it contains only one cluster. The Brodinova et al. \cite{brodinova2017robust} generator was used to generate high-dimensional data sets consisting of informative and non-informative features. No noise injection (attributes with noise contamination) was considered in the current study. To avoid situations of overlapping clusters the minimum Euclidean distance between any two points in different clusters was set to 3 and data sets violating this rule were re-generated. A summary of the models can be found in Table \ref{sdatasets3}. Next we use our own synthetic data sets models consisted of clusters with mixed properties. We will refer to these models as \textit{mixed} (refer to Figure \ref{variousModels} for a sample visualization of these models). 

\begin{table}
	\centering
	\caption{\textbf{Gap \cite{tibshirani2001estimating} and weighted gap statistic \cite{yan2007determining} data sets models.} Points: the number of data points per cluster, \textit{or} indicates that a random number was selected among the specified numbers for each cluster, \textit{to} indicates that a random number was selected between the specified numbers for each cluster; p: number of features or attributes of the data set (dimensions); K: number of generated clusters. Gaussian models: clusters of low dimensionality generated from Gaussian distributions. 10-D Gaussian models: clusters of higher dimensionality generated from Gaussian distributions. Elongated models: clusters generated by adding Gaussian noise across lines. Unbalanced model: data sets containing Gaussian clusters of very different sizes, exponential: non-Gaussain clusters generated from the exponential distribution. For more information about the parameters used for the Gaussians refer to the relevant studies and the supplementary material. Visualization (when possible) of a data set from each model is available in Figure \ref{variousModels}.} \label{gdatasets}		
	\begin{tabular}{ccccccccc}
		\cline{1-4} \cline{6-9}
		\multicolumn{1}{|c|}{\textbf{\begin{tabular}[c]{@{}c@{}}Gaussian\\ models\end{tabular}}}        & \multicolumn{1}{c|}{\textbf{Points}} & \multicolumn{1}{c|}{\textbf{p}} & \multicolumn{1}{c|}{\textbf{K}} & \multicolumn{1}{c|}{} & \multicolumn{1}{c|}{\textbf{\begin{tabular}[c]{@{}c@{}}Unbalanced\\ model\end{tabular}}}       & \multicolumn{1}{c|}{\textbf{Points}} & \multicolumn{1}{c|}{\textbf{p}} & \multicolumn{1}{c|}{\textbf{K}} \\ \cline{1-4} \cline{6-9} 
		\multicolumn{1}{|c|}{\begin{tabular}[c]{@{}c@{}}gap model\\ (gap 2)\end{tabular}}               & \multicolumn{1}{c|}{25,25,50}        & \multicolumn{1}{c|}{2}          & \multicolumn{1}{c|}{3}          & \multicolumn{1}{c|}{} & \multicolumn{1}{c|}{\begin{tabular}[c]{@{}c@{}}weighted gap\\ model 2\\ (wgap 2)\end{tabular}} & \multicolumn{1}{c|}{100,15}          & \multicolumn{1}{c|}{2}          & \multicolumn{1}{c|}{2}          \\ \cline{1-4} \cline{6-9} 
		\multicolumn{1}{|c|}{\begin{tabular}[c]{@{}c@{}}gap model 3\\ (gap 3)\end{tabular}}             & \multicolumn{1}{c|}{25 or 50}        & \multicolumn{1}{c|}{3}          & \multicolumn{1}{c|}{4}          &                       &                                                                                                &                                      &                                 &                                 \\ \cline{1-4} \cline{6-9} 
		\multicolumn{1}{|c|}{\begin{tabular}[c]{@{}c@{}}weighted gap\\ model 1\\ (wgap 1)\end{tabular}} & \multicolumn{1}{c|}{25 to 50}        & \multicolumn{1}{c|}{2}          & \multicolumn{1}{c|}{6}          & \multicolumn{1}{c|}{} & \multicolumn{1}{c|}{\textbf{\begin{tabular}[c]{@{}c@{}}Exponential\\ model\end{tabular}}}      & \multicolumn{1}{c|}{\textbf{Points}} & \multicolumn{1}{c|}{\textbf{p}} & \multicolumn{1}{c|}{\textbf{K}} \\ \cline{1-4} \cline{6-9} 
		\multicolumn{1}{|c|}{\begin{tabular}[c]{@{}c@{}}weighted gap\\ model 6\\ (wgap 6)\end{tabular}} & \multicolumn{1}{c|}{50 each}         & \multicolumn{1}{c|}{2}          & \multicolumn{1}{c|}{6}          & \multicolumn{1}{c|}{} & \multicolumn{1}{c|}{\begin{tabular}[c]{@{}c@{}}weighted gap\\ model 3\\ (wgap 3)\end{tabular}} & \multicolumn{1}{c|}{50 each}         & \multicolumn{1}{c|}{2}          & \multicolumn{1}{c|}{4}          \\ \cline{1-4} \cline{6-9} 
		&                                      &                                 &                                 & \multicolumn{1}{l}{}  & \multicolumn{1}{l}{}                                                                           & \multicolumn{1}{l}{}                 & \multicolumn{1}{l}{}            & \multicolumn{1}{l}{}            \\ \cline{1-4} \cline{6-9} 
		\multicolumn{1}{|c|}{\textbf{\begin{tabular}[c]{@{}c@{}}10-D Gaussian\\ models\end{tabular}}}   & \multicolumn{1}{c|}{\textbf{Points}} & \multicolumn{1}{c|}{\textbf{p}} & \multicolumn{1}{c|}{\textbf{K}} & \multicolumn{1}{c|}{} & \multicolumn{1}{c|}{\textbf{\begin{tabular}[c]{@{}c@{}}Elongated\\ models\end{tabular}}}       & \multicolumn{1}{c|}{\textbf{Points}} & \multicolumn{1}{c|}{\textbf{p}} & \multicolumn{1}{c|}{\textbf{K}} \\ \cline{1-4} \cline{6-9} 
		\multicolumn{1}{|c|}{\begin{tabular}[c]{@{}c@{}}gap model 4\\ (gap 4)\end{tabular}}             & \multicolumn{1}{c|}{25 or 50}        & \multicolumn{1}{c|}{10}         & \multicolumn{1}{c|}{2}          & \multicolumn{1}{c|}{} & \multicolumn{1}{c|}{\begin{tabular}[c]{@{}c@{}}gap model 5\\ (gap 5)\end{tabular}}             & \multicolumn{1}{c|}{100 each}        & \multicolumn{1}{c|}{3}          & \multicolumn{1}{c|}{2}          \\ \cline{1-4} \cline{6-9} 
		\multicolumn{1}{|c|}{\begin{tabular}[c]{@{}c@{}}weighted gap\\ model 5\\ (wgap 5)\end{tabular}} & \multicolumn{1}{c|}{25 to 50}        & \multicolumn{1}{c|}{10}         & \multicolumn{1}{c|}{2}          & \multicolumn{1}{c|}{} & \multicolumn{1}{c|}{\begin{tabular}[c]{@{}c@{}}weighted gap\\ model 4\\ (wgap 4)\end{tabular}} & \multicolumn{1}{c|}{100 each}        & \multicolumn{1}{c|}{2}          & \multicolumn{1}{c|}{2}          \\ \cline{1-4} \cline{6-9} 
	\end{tabular}
\end{table}

\begin{table}
	\centering
	\caption{\textbf{Brodinova model generator \cite{brodinova2017robust}.} The minimum allowed Euclidean distance between two data points in different clusters was set to 3 and no noise injection was considered. Name: name of the model; Points: the total number of data points in the data set; p: number of features or attributes of the data set, \textit{Informative (+)} indicates attributes that are required to describe the data set while \textit{Non-informative (-)} indicates variables that should be ignored; K: number of generated clusters. These models are creating high-dimensional Gaussian clusters of different shapes using two different distributions, one for the informative and one for the uninformative variables. \textit{Left table:} Each cluster contains 40 data points. The parameters of these models are selected to test the performance of the clustering algorithm in data sets with different degrees of informative and/or uninformative features \textit{Right table:} The first four models create higher-dimensional balanced clusters (clusters of equal sizes) and the last two higher-dimensional unbalanced clusters each with number of points randomly selected between 50 to 100. The parameters of these models are selected to test the performance of the clustering algorithm in balanced and unbalanced data sets of higher dimensionality with increasing number of clusters. The input space was selected to be sparse, i.e. a few hundred points in 1000 or 1500 dimensions to avoid the slow computation of the Kaufman algorithm.}
	\label{sdatasets3}	
	\begin{multicols}{2}
		\begin{tabular}{|l|c|c|c|c|}
			\hline
			\multicolumn{1}{|c|}{\multirow{2}{*}{\textbf{Name}}} & \multirow{2}{*}{\textbf{Points}} & \multicolumn{2}{c|}{\textbf{p}}                                         & \multirow{2}{*}{\textbf{K}} \\ \cline{3-4}
			\multicolumn{1}{|c|}{}                      &                              & \multicolumn{1}{c|}{+} & \multicolumn{1}{c|}{-} &                           \\ \hline
			brod 1                                    & \hphantom{xx}120\hphantom{xx}                          & \hphantom{x}20\hphantom{x}                               & 0                                    & 3                         \\ \hline
			brod 2                                    & 400                          & 20                               & 0                                    & 10                        \\ \hline
			brod 3                                    & 120                          & 15                               & 5                                    & 3                         \\ \hline
			brod 4                                    & 400                          & 15                               & 5                                    & 10                        \\ \hline
			brod 5                                    & 120                          & 10                               & 10                                   & 3                         \\ \hline
			brod 6                                    & 400                          & 10                               & 10                                   & 10                        \\ \hline
		\end{tabular}
		\vfill
		\begin{tabular}{|l|c|c|c|c|}
			\hline
			\multicolumn{1}{|c|}{\multirow{2}{*}{\textbf{Name}}} & \multirow{2}{*}{\textbf{Points}} & \multicolumn{2}{c|}{\textbf{p}}                                         & \multirow{2}{*}{\textbf{K}} \\ \cline{3-4}
			\multicolumn{1}{|c|}{}                      &                              & \multicolumn{1}{c|}{+} & \multicolumn{1}{c|}{-} &                           \\ \hline
			brod 7                                    & 120                          & 1000                               & 0                                    & 3                         \\ \hline
			brod 8                                    & 400                          & 1000                               & 0                                    & 10                       \\ \hline
			brod 9                                    & 400                          & 1500                               & 0                                    & 10                       \\ \hline		
			brod 10                                    & 1250                          & 1500                               & 0                                    & 50                         \\ \hline
			brod 11                                    & 50 to 100                         & 1000                               & 0                                    & 3                        \\ \hline
			brod 12                                    & 50 to 100                          & 1000                               & 0                                   & 10                         \\ \hline
		\end{tabular}
	\end{multicols}
\end{table}

\begin{itemize}
	\item[$\bullet$] \textit{model 1} generates 3-dimensional clusters, 1 spherical and 2 elongated. The spherical cluster is an 80 points Gaussian cluster at the origin with standard deviation of 0.1. The two elongated clusters have 100 points each and are generated as follows: $x_1=x_2=x_3=t$ with $t$ taking 100 equally spaced values from -1 to 1. We then add Gaussian noise with standard deviation of 0.3 to each dimension. The second dimension of the first elongated cluster was shifted by 2 from the centre of the spherical cluster. Similarly the second dimension of the second elongated cluster was shifted by -2 and the first dimension was rotated by $180^0$.
	\item[$\bullet$] \textit{model 2} generates 3-dimensional non-Gaussian and normal clusters. It generates: (a) a cluster from an exponential distribution with rate of 1 and truncated at $[-1 \hphantom{x} 1]$ containing 80 points, (b) a cluster from an exponential distribution with rate of 1 and and truncated at $[2 \hphantom{x} 3]$ with 100 points, (c) a Gaussian cluster of 80 points with mean $[0.5,2.5,2.5]$ and standard deviation of 0.1 in every dimension and (d) a Gaussian cluster of 100 points with mean $[2.5,0.5,0.5]$ and standard deviation of 0.2 in every dimension.
	\item[$\bullet$] \textit{model 3} generates 3-dimensional Gaussian clusters with different standard deviations. The first cluster has 80 points with mean at the origin and standard deviation of 0.1 on each dimension. The second cluster has of 100 points with mean $[2,0,0]$ and standard deviation of 0.2 on each dimension. The third cluster has of 120 points with mean $[0,2,0]$ and standard deviation of 0.3 on each dimension. The forth cluster consists of 140 points with mean $[0,0,2]$ and standard deviation of 0.4 on each dimension.
	\item[$\bullet$] \textit{model 4} generates 3-dimensional mixed Gaussian clusters. The first cluster consists of 80 points with mean $[0,0,0]$ and standard deviations $[0.1,0.1,0.2]$. The second cluster consists of 100 points with mean $[2,0,0]$ and standard deviations $[0.1,0.2,0.3]$. The third cluster consists of 120 points with mean $[0,2,0]$ and standard deviations $[0.2,0.4,0.6]$. The forth cluster consists of 140 points with mean $[0,0,2]$ and standard deviations $[1.0,0.1,0.1]$.
\end{itemize}

We also consider the \textit{S-sets} \cite{Ssets} and the \textit{A-sets} \cite{Asets} obtained from the ``clustering basic benchmark'' which was used in the studies of \cite{franti2018k,franti2019much}. The aforementioned studies were dedicated to the K-Means properties, advantages and disadvantages and assessed used various synthetic data sets. Both models contains 2-dimensional data; \textit{S-sets} contains 4 data sets with 5000 data points distributed among 15 Gaussian clusters with different degree of clustering overlap \cite{Ssets} and \textit{A-sets} contains 3 data sets with 20, 35 and 50 clusters and 150 data points per cluster \cite{Asets}. For more information about these data sets refer to the relevant studies. Finally we considered a selection of data sets from the UCI repository \cite{asuncion2007uci}: Iris, Ionosphere, Wine, Breast Cancer, Glass and Yeast. More information about these data sets are shown on Table \ref{realsets}.

\begin{table}[]
	\centering
	\caption{\textbf{Real data sets from the UCI repository \cite{asuncion2007uci}.} Points: the number of data points per cluster; p: number of features or attributes of the data set (dimensions); K: number of generated clusters.}
	\label{realsets}	
	\begin{tabular}{|l|c|c|c|}
		\hline
		\multicolumn{1}{|c|}{\textbf{Name}} & \textbf{Points}                                                                & \textbf{p} & \textbf{K} \\ \hline
		Iris                                & 50,50,50                                      & 4          & 3          \\ \hline
		Ionosphere                          & 225,126                                                                        & 34         & 2          \\ \hline
		Wine                                & 59,71,48                                                                       & 13         & 3          \\ \hline
		Breast Cancer                       & 444,239                                                                        & 9          & 2          \\ \hline
		Glass                               & \begin{tabular}[c]{@{}c@{}}70,76,17,\\ 19,9,29\end{tabular}                    & 9          & 6          \\ \hline
		Yeast                               & \begin{tabular}[c]{@{}c@{}}463,5,35,44,\\ 51,163,244,\\ 429,20,30\end{tabular} & 8          & 10         \\ \hline
	\end{tabular}
\end{table}

\section{Results}

We test the performance of the K-Means variations, Lloyd's \cite{jain2010data} Hartigan-Wong's \cite{hartigan1979algorithm,hartigan1975clustering} and K-Medians \cite{charu2013data} initialised using the eight different clustering initialisation methods named: Random \cite{macqueen1967some}, K-Means++ \cite{arthur2007k}, Maximin(S) \cite{gonzalez1985clustering}, ROBIN(S) \cite{brodinova2017robust}, Kaufman \cite{kaufman2009finding}, ROBIN(D) \cite{al2009robust}, DK-Means++ \cite{nidheesh2017enhanced} and Maximin(D) \cite{katsavounidis1994new}. For the ROBIN variations the $mp$ parameter specifying the number of neighbor data points was set to $10$ as in the original study \cite{al2009robust}. For the Hartigan-Wong's algorithm NAG's implementation was used \cite{NAGlib}.

We conceptually consider a ``sophistication'' scale for the initialisation methods based not only on their execution time but also on the complexity of their underlying operators. For example DK-Means++ and ROBIN would be considered more sophisticated than Kaufman since they incorporate more advanced statistics while Kaufman uses only distances and still has a complexity of $O(N^2)$. Our scale is as follows: Random $<$ K-Means++ $<$ Maximin $<$ Kaufman $<$ ROBIN $<$ DK-Means++. 

In our experiments we use the synthetic data sets models from the studies of gap statistic \cite{tibshirani2001estimating} and weighted gap statistic \cite{yan2007determining} (refer to Table \ref{gdatasets}, 10 sets in total), Brodinova \cite{brodinova2017robust} (refer to Table \ref{sdatasets3}, 12 sets in total) and other four custom data sets models (refer to Methods and Figure \ref{variousModels}, 4 sets in total). From each model we generated 40 data sets and for each data set the stochastic methods were executed 50 times. These numbers were selected to provide a good statistical sample. We also use the ``clustering data sets'' (\textit{S-sets} \cite{Ssets} and \textit{A-sets} \cite{Asets}) from the studies of \cite{franti2018k,franti2019much} and real-world data sets from the UCI repository \cite{asuncion2007uci}: Iris, Ionosphere, Wine, Breast Cancer, Glass and Yeast (see Table \ref{realsets}). For each of these data sets we use the same set up of executing the stochastic methods 50 times. 

For all our hypothesis testing on the data set models we used the Paired Samples Wilcoxon Test, a non-parametric alternative to paired t-test, For the outcome of the test we generally use the following symbols for the level of significance, \textit{$*$} for p-value $<$ 0.05; \textit{$**$} for p-value $<$ 0.01; \textit{$***$} for p-value $<$ 0.001; \textit{$****$} for p-value $<$ 0.0001. We evaluated the monotonic relationship of Silhouette index and Purity via a large sample of clustering results on the multiple executions of the methods across all our data sets (20000 cases). Using Spearman's rank correlation coefficient, we confirmed that Purity and Silhouette have a strong monotonic relation (Spearman's Rho $0.97$). As a side note, we also considered Distortion \cite{al2009robust} as an unsupervised index but found the monotonic relationship with Purity weaker (Spearman's Rho $0.65$).

\begin{table}[h]
	\caption{\textbf{Average performance comparison among K-Means variations using simple stochastic and more sophisticated deterministic initialisation methods.} We compare two K-Means variations (Hartigan-Wong's K-Means: HW, Lloyd's K-Means: Ll and K-Medians: KMed) initialised with the same method on 26 occasions (10 gap and weighted gap, 12 Brodinova and 4 mixed models). To calculate performance, we averaged the Purity index across the 50 initial conditions and 40 data sets for each model and the comparison is based on the times that there was significant difference between the two algorithms.} \label{TalgosMean}
	\centering
	\begin{tabular}{|l|c|c|c|c|}
		\hline
		\multicolumn{1}{|c|}{} &
		&
		\multicolumn{3}{c|}{\textbf{\begin{tabular}[c]{@{}c@{}}Significantly better\\ average performance\end{tabular}}} \\ \cline{3-5} 
		\multicolumn{1}{|c|}{\multirow{-2}{*}{\textbf{\begin{tabular}[c]{@{}c@{}}Initialization\\ method\end{tabular}}}} &
		\multirow{-2}{*}{\textbf{\begin{tabular}[c]{@{}c@{}}Total number\\ of instances\end{tabular}}} &
		\textbf{HW vs Ll} &
		\textbf{HW vs KMed} &
		\textbf{Ll vs KMed} \\ \hline
		\textbf{Random}     & 26 & 13 vs 0 & 18 vs 4 & 6 vs 4 \\ \hline
		\textbf{K-Means++}  & 26 & 10 vs 0 & 13 vs 3 & 2 vs 5 \\ \hline
		\rowcolor[HTML]{000000} 
		{\color[HTML]{FFFFFF} \textbf{Total}} &
		{\color[HTML]{FFFFFF} \textbf{52}} &
		{\color[HTML]{FFFFFF} \textbf{23 vs 0}} &
		{\color[HTML]{FFFFFF} \textbf{39 vs 12}} &
		{\color[HTML]{FFFFFF} \textbf{8 vs 9}} \\ \hline
		\textbf{Kaufman}    & 26 & 1 vs 1  & 2 vs 6  & 1 vs 6 \\ \hline
		\textbf{DK-Means++} & 26 & 1 vs 0  & 2 vs 4  & 1 vs 5 \\ \hline
		\rowcolor[HTML]{000000} 
		{\color[HTML]{FFFFFF} \textbf{Total}} &
		{\color[HTML]{FFFFFF} \textbf{52}} &
		{\color[HTML]{FFFFFF} \textbf{2 vs 1}} &
		{\color[HTML]{FFFFFF} \textbf{4 vs 10}} &
		{\color[HTML]{FFFFFF} \textbf{2 vs 11}} \\ \hline
	\end{tabular}
\end{table}

\subsection{Comparison on the average performance among stochastic and among deterministic methods}

We assess on average the performance of stochastic methods as well as the performance of deterministic methods. We calculate the average performance of stochastic methods on 50 different runs across 40 different data sets for each one of our 26 models (10 gap and weighted gap, 12 Brodinova, 4 mixed models). Deterministic methods were executed once on the 40 data sets. 

\begin{figure}
	\centering
	\includegraphics[width=0.9\linewidth]{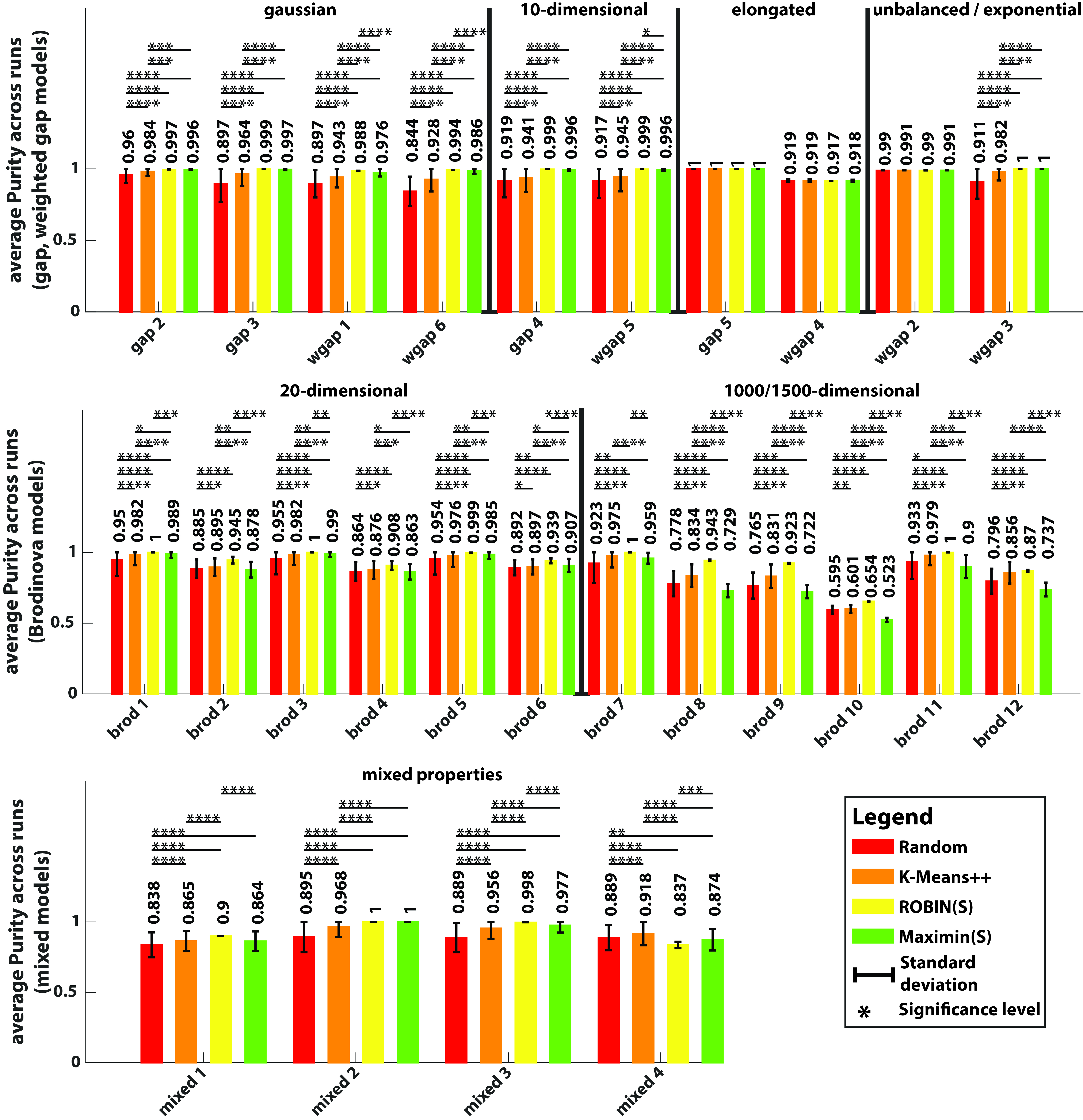}
	
	\begin{tabular}{|l|c|c|c|c|}
		\hline
		\multicolumn{1}{|c|}{}                                            &                                                                                                & \multicolumn{3}{c|}{\textbf{\begin{tabular}[c]{@{}c@{}}Significantly better\\ average performance \end{tabular}}}                                                    \\ \cline{3-5} 
		\multicolumn{1}{|c|}{\multirow{-2}{*}{\textbf{\begin{tabular}[c]{@{}c@{}}Initialization\\ method\end{tabular}}}}                    & \multirow{-2}{*}{\textbf{\begin{tabular}[c]{@{}c@{}}Total number\\ of instances\end{tabular}}} & \textbf{HW}                                                 & \textbf{Ll}                                                 & \textbf{KMed}                                               \\ \hline
		\textbf{Random vs K-Means++}                                                                                                        & 26                                                                                             & 0 vs 23 & 0 vs 23 & 0 vs 24  \\ \hline
		\textbf{Random vs ROBIN(S)}                                                                                                         & 26                                                                                             & 1 vs 22 & 3 vs 22 & 2 vs 23 \\ \hline
		\textbf{Random vs Maximin(S)}                                                                                                       & 26                                                                                             & 6 vs 15 & 6 vs 16 & 6 vs 16 \\ \hline
		\textbf{K-Means++ vs ROBIN(S)}                                                                                                      & 26                                                                                             & 1 vs 21 & 3 vs 21 & 2 vs 22 \\ \hline
		\textbf{K-Means++ vs Maximin(S)}                                                                                                    & 26                                                                                             & 8 vs 13 & 9 vs 14 & 7 vs 13 \\ \hline
		\textbf{ROBIN(S) vs Maximin(S)}                                                                                                     & 26                                                                                             & 17 vs 1 & 18 vs 3 & 19 vs 1 \\ \hline
	\end{tabular}
	\caption{\textbf{Comparison on the average performance among stochastic initialisation methods.} Each plot shows the performance of the Hartigan-Wong's K-Means clustering solution using the Purity index (y-axis) on different data sets models (x-axis) and initialized with different stochastic methods. To calculate performance, we averaged the Purity index across the 50 initial conditions and 40 data sets for each model (gap, weighted gap, Brodinova and mixed). The errorbars are showing the (average) standard deviation across the 40 data sets. Solid lines on any two bars underline the level of significant difference between the corresponding methods. The accompanied Table below the figure shows a summary of the comparisons through all the K-Means variations (Hartigan-Wong's K-Means (HW), Lloyd's K-Means (Ll) and K-Medians (KMed)) where there is a significant performance difference between the compared methods.} \label{fig1}  
\end{figure}

\begin{figure}
	\centering
	\includegraphics[width=0.9\linewidth]{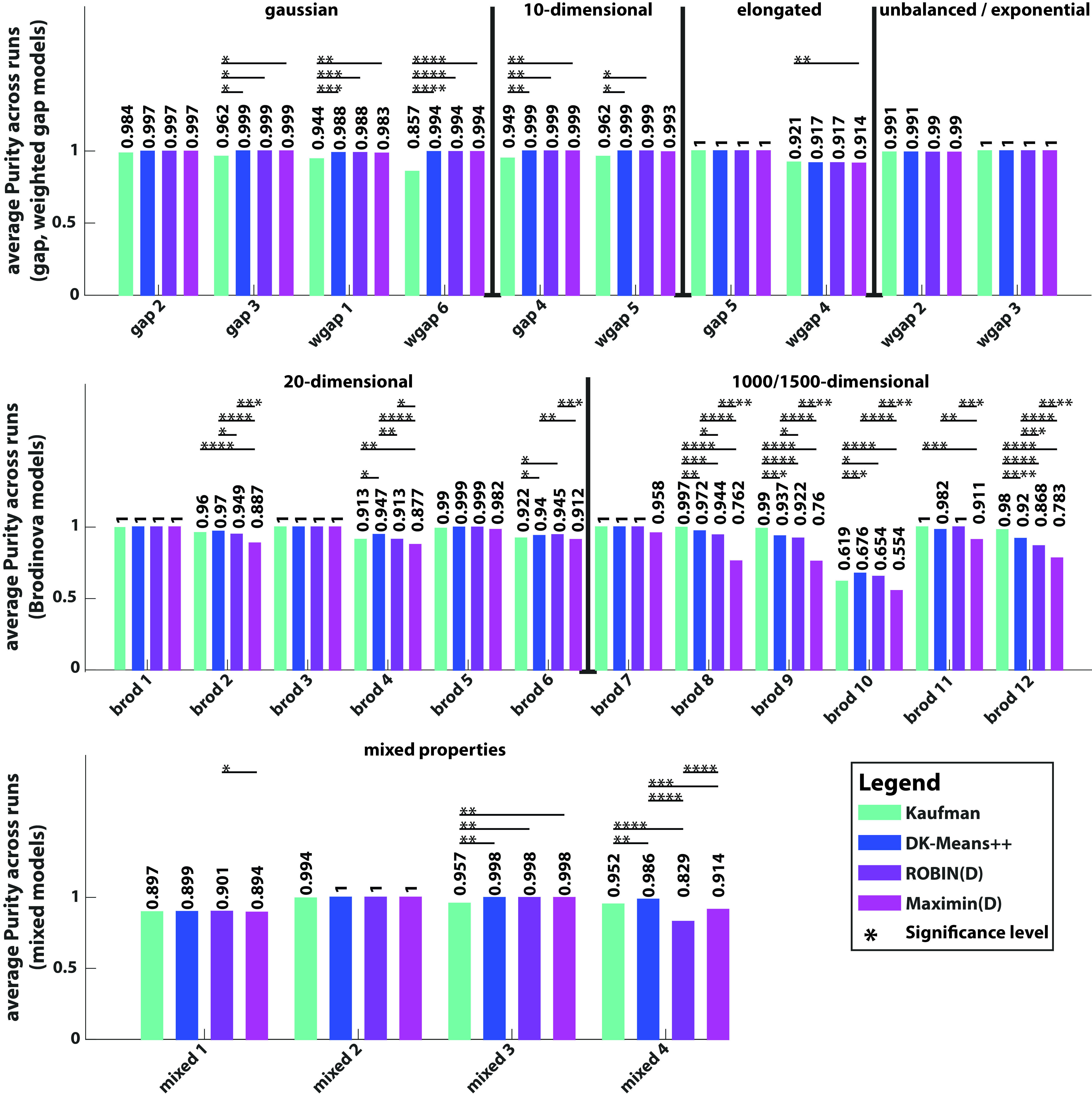}
	
	\begin{tabular}{|l|c|c|c|c|}
		\hline
		\multicolumn{1}{|c|}{}                                            &                                                                                                & \multicolumn{3}{c|}{\textbf{\begin{tabular}[c]{@{}c@{}}Significantly better\\ average performance \end{tabular}}}                                                    \\ \cline{3-5} 
		\multicolumn{1}{|c|}{\multirow{-2}{*}{\textbf{\begin{tabular}[c]{@{}c@{}}Initialization\\ method\end{tabular}}}}                    & \multirow{-2}{*}{\textbf{\begin{tabular}[c]{@{}c@{}}Total number\\ of instances\end{tabular}}} & \textbf{HW}                                                 & \textbf{Ll}                                                 & \textbf{KMed}                                               \\ \hline
		\textbf{Kaufman vs DK-Means++}                                                                                                        & 26                                                                                             & 3 vs 10 & 3 vs 9 & 4 vs 8  \\ \hline
		\textbf{Kaufman vs ROBIN(D)}                                                                                                         & 26                                                                                             & 4 vs 8 & 6 vs 9 & 4 vs 6 \\ \hline
		\textbf{Kaufman vs Maximin(D)}                                                                                                       & 26                                                                                             & 8 vs 5 & 11 vs 5 & 9 vs 5 \\ \hline
		\textbf{DK-Means++ vs ROBIN(D)}                                                                                                      & 26                                                                                             & 6 vs 0 & 7 vs 0 & 6 vs 1 \\ \hline
		\textbf{DK-Means++ vs Maximin(D)}                                                                                                    & 26                                                                                             & 9 vs 0 & 11 vs 0 & 10 vs 1 \\ \hline
		\textbf{ROBIN(D) vs Maximin(D)}                                                                                                     & 26                                                                                             & 9 vs 1 & 10 vs 1 & 9 vs 1 \\ \hline
	\end{tabular}
	\caption{\textbf{Comparison on the performance among deterministic initialisation methods.} Each plot shows the performance of the Hartigan-Wong's K-Means clustering solution using the Purity index (y-axis) on different data sets models (x-axis) and initialized with different deterministic methods. To calculate performance, we averaged the Purity index across the 40 data sets for each model (gap, weighted gap, Brodinova and mixed). Solid lines on any two bars underline the level of significant difference between the corresponding methods (cases of no significant differences are not showing). The accompanied Table below the figure shows a summary of the comparisons through all the K-Means variations (Hartigan-Wong's K-Means (HW), Lloyd's K-Means (Ll) and K-Medians (KMed)) where there is a significant performance difference between the compared methods.} \label{fig2}  
\end{figure}

Based on Figure \ref{fig1} the average performance of K-Means variations increases by using more sophisticated initialisation methods and ROBIN(S) initialisation provides the best average performance followed by Maximin(S) and K-Means++ while Random initialisation results in the poorest performance. For the deterministic methods, shown on Figure \ref{fig2}, we observe again that the average performance of K-Means variations increases by using more sophisticated initialisation methods. DK-Means++ achieved the best average performance followed by ROBIN(D) and then by Kaufman and Maximin(D). Finally we wanted to assess if more sophisticated initialization methods alleviate the need for complex clustering. For this reason we performed comparisons among the K-Means variations initialised with either Random and K-Means++ or Kaufman and DK-Means++ methods. Maximin and ROBIN have both stochastic and deterministic variations of equal sophistication thus we excluded them. As shown in Table \ref{TalgosMean}, deterministic methods that are more sophisticated (as per our definition) reduce performance differences among the different variants of the K-Means algorithms.

\subsection{Comparison of the average performance between stochastic and deterministic methods}

\begin{figure}[h]
	\centering
	\includegraphics[width=\linewidth]{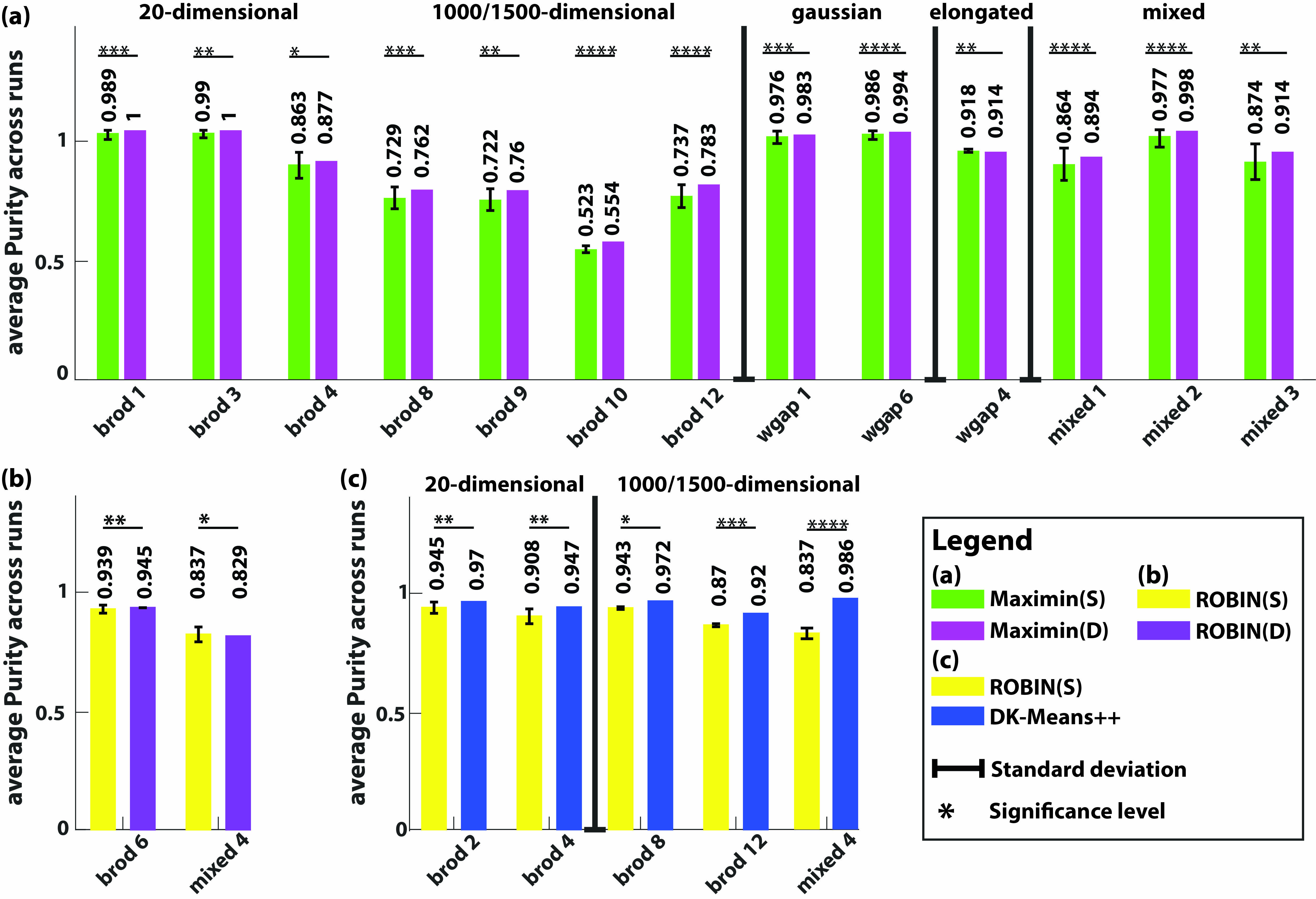}
	
	\begin{tabular}{|l|c|c|c|c|}
		\hline
		\multicolumn{1}{|c|}{\multirow{2}{*}{\textbf{\begin{tabular}[c]{@{}c@{}}Initialization\\ method\end{tabular}}}} & \multirow{2}{*}{\textbf{\begin{tabular}[c]{@{}c@{}}Total number\\ of instances\end{tabular}}} & \multicolumn{3}{c|}{\textbf{\begin{tabular}[c]{@{}c@{}}Significantly better\\ average performance\end{tabular}}}                                              \\ \cline{3-5} 
		\multicolumn{1}{|c|}{}                                                                                          &                                                                                               & \textbf{HW}                                               & \textbf{Ll}                                               & \textbf{KMed}                                             \\ \hline
		\textbf{Maximin(S) vs Maximin(D)}                                                                                   & 26                                                                                            & 1 vs 12 & 3 vs 12 & 0 vs 13 \\ \hline
		\textbf{ROBIN(S) vs ROBIN(D)}                                                                                   & 26                                                                                            & 1 vs 1 & 1 vs 2 & 1 vs 0 \\ \hline
		\textbf{ROBIN(S) vs DK-Means++}                                                                                 & 26                                                                                            & 0 vs 5 & 0 vs 7 & 1 vs 5 \\ \hline
	\end{tabular}
	\caption{\textbf{Comparisons on the average performance between stochastic and deterministic methods.} Each plot shows the performance of the Hartigan-Wong's K-Means clustering solution using the Purity index (y-axis) on different data sets models (x-axis) and initialized with different stochastic methods (only the cases where significant difference was present are shown). To calculate performance, we averaged the Purity index across the 50 initial conditions and 40 data sets for each model (gap, weighted gap and Brodinova). The errorbars (on the stochastic methods only) are showing the average standard deviation across the 40 data sets. Solid lines on any two bars underline the level of significant difference between the corresponding methods (cases of no significant differences are not showing). The accompanied Table below the figure shows a summary of the comparisons through all the K-Means variations (Hartigan-Wong's K-Means (HW), Lloyd's K-Means (Ll) and K-Medians (KMed)) where there is a significant performance difference between the compared methods.} \label{fig3}  
\end{figure}

Following on from our previous conclusions, we wanted to assess if overall deterministic methods provide on average better performance than stochastic methods. For this reason we compared the stochastic and deterministic variations of Maximin and ROBIN as well as the best stochastic performer (ROBIN(S) see Figure \ref{fig1}) and the best deterministic performer (DK-Means++ see Figure \ref{fig2}). Based on the results in Figure \ref{fig3}: (a) Maximin(D) is on average better than Maximin(S); (b) ROBIN(D) and ROBIN(S) are on average equivalent; (c) DK-Means++ is better than ROBIN(S).

\subsection{Comparison of the maximum performance across multiple runs of stochastic and deterministic methods}

Next, we wanted to compare the stochastic and the deterministic methods but based on the maximum performance that the former can achieve on multiple repetitions. We run each of the stochastic methods 50 times and select the best outcome based on the silhouette index. We then report its corresponding value according to the purity index. We expect that due to the many repetitions, stochastic methods can find different local minima and potentially result in a better performance at the cost of multiple repetitions.

\begin{figure}
	\centering
	\includegraphics[width=0.9\linewidth]{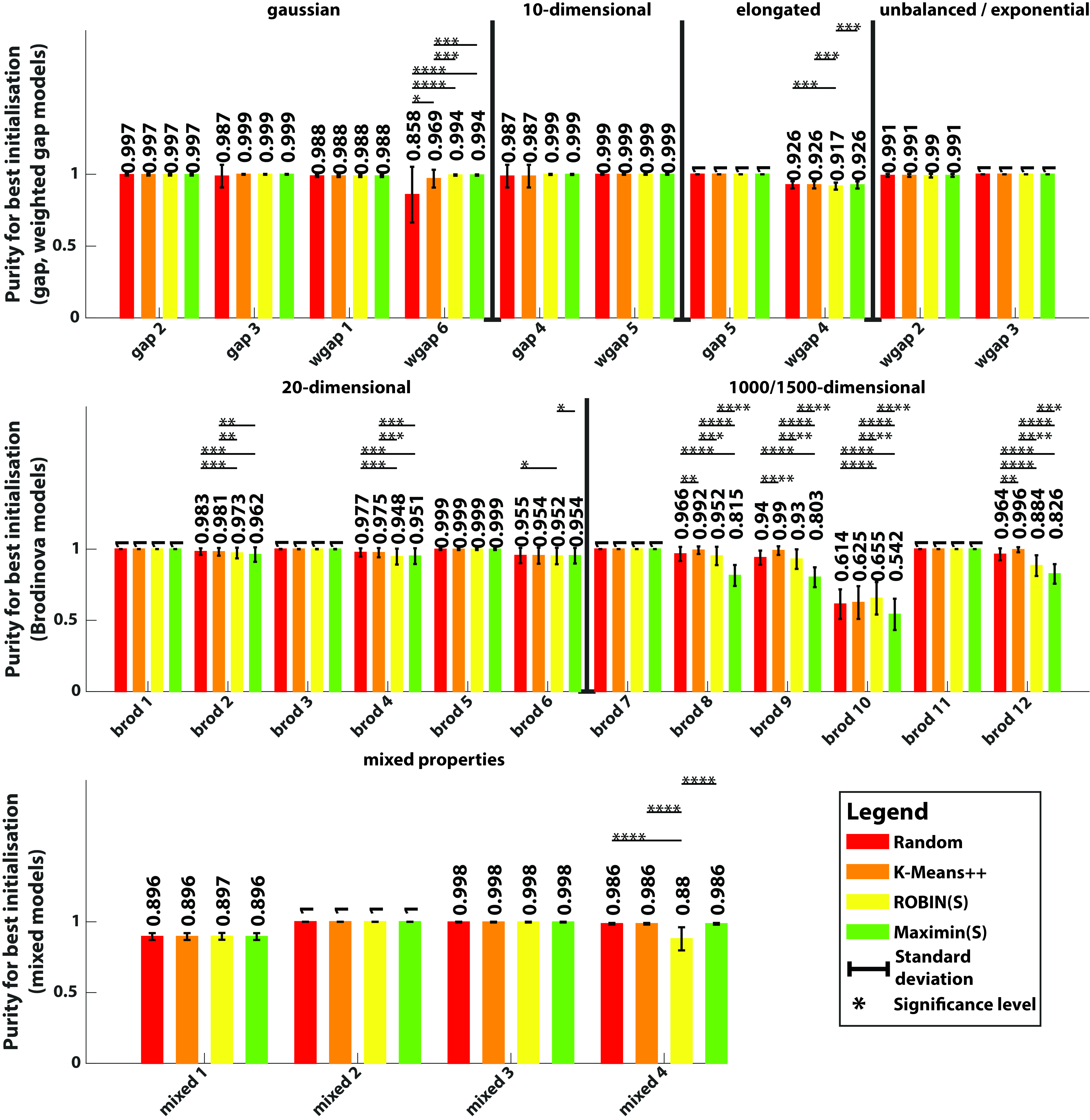}
	\begin{tabular}{|l|c|c|c|c|}
		\hline
		\multicolumn{1}{|c|}{\multirow{2}{*}{\begin{tabular}[c]{@{}c@{}}Initialization\\ method\end{tabular}}} & \multirow{2}{*}{\begin{tabular}[c]{@{}c@{}}Total number\\ of instances\end{tabular}} & \multicolumn{3}{c|}{\begin{tabular}[c]{@{}c@{}}Significantly better\\ maximum performance\\ Purity for best Silhouette\end{tabular}} \\ \cline{3-5} 
		\multicolumn{1}{|c|}{}                                                                                 &                                                                                      & HW                                         & Ll                                        & KMed                                      \\ \hline
		Random vs K-Means++                                                                                    & 26                                                                                   & 0 vs 4                                     & 0 vs 4                                    & 1 vs 4                                    \\ \hline
		Random vs ROBIN(S)                                                                                     & 26                                                                                   & 6 vs 3                                     & 8 vs 2                                    & 6 vs 2                                    \\ \hline
		Random vs Maximin(S)                                                                                   & 26                                                                                   & 6 vs 1                                     & 9 vs 1                                    & 9 vs 1                                    \\ \hline
		K-Means++ vs ROBIN(S)                                                                                  & 26                                                                                   & 7 vs 2                                     & 9 vs 2                                    & 8 vs 2                                    \\ \hline
		K-Means++ vs Maximin(S)                                                                                & 26                                                                                   & 6 vs 1                                     & 9 vs 1                                    & 9 vs 1                                    \\ \hline
		ROBIN(S) vs Maximin(S)                                                                                 & 26                                                                                   & 4 vs 3                                     & 4 vs 4                                    & 4 vs 2                                    \\ \hline
	\end{tabular}
	\caption{\textbf{Comparison on the maximum performance among stochastic initialisation methods.} 
		Each plot shows the performance of the Hartigan-Wong's K-Means clustering solution using the purity corresponding to the best silhouette score achieved within 50 different executions (y-axis) on different data sets models (x-axis) and initialized with different stochastic methods. Purity for best silhouette score was averaged over the 40 data sets for each model (gap, weighted gap and Brodinova)
		The errorbars are showing the standard deviation across the 40 data sets. Solid lines on any two bars underline the level of significant difference between the corresponding methods. The accompanied Table below the figure shows a summary of the comparisons through all the K-Means variations (Hartigan-Wong's K-Means (HW), Lloyd's K-Means (Ll) and K-Medians (KMed)) where there is a significant performance difference between the compared methods.} \label{fig4}  
\end{figure}

Firstly, we repeat the comparison among the different stochastic methods but based on the maximum performance that they can achieve. Figure \ref{fig4} shows the relevant results and, opposite to our observations on the average performance, stochastic methods have higher chances of obtaining a better clustering result with multiple execution. K-Means++ is the best method followed by Random while ROBIN(S) and Maximin(S) have almost similar performance.

\begin{figure}
	\centering
	\includegraphics[width=\linewidth]{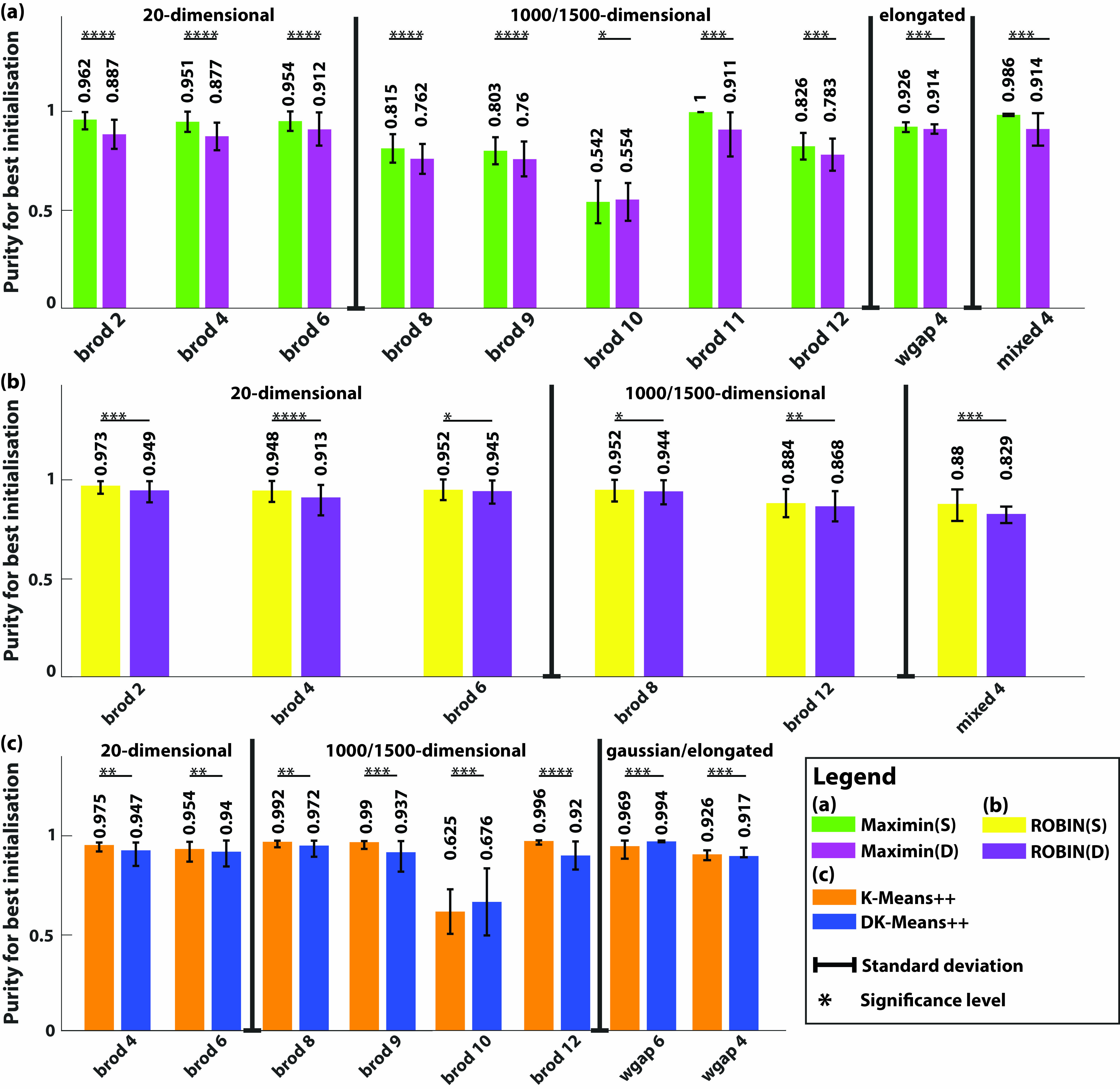}
	\begin{tabular}{|l|c|c|c|c|}
		\hline
		\multicolumn{1}{|c|}{\multirow{2}{*}{\textbf{\begin{tabular}[c]{@{}c@{}}Initialization\\ method\end{tabular}}}} & \multirow{2}{*}{\textbf{\begin{tabular}[c]{@{}c@{}}Total number\\ of instances\end{tabular}}} & \multicolumn{3}{c|}{\textbf{\begin{tabular}[c]{@{}c@{}}Significant better\\ average performance\end{tabular}}}                                              \\ \cline{3-5} 
		\multicolumn{1}{|c|}{}                                                                                          &                                                                                               & \textbf{HW}                                               & \textbf{Ll}                                               & \textbf{KMed}                                             \\ \hline
		\textbf{Maximin(S) vs Maximin(D)}                                                                                   & 26                                                                                            & 9 vs 1 & 11 vs 1 & 10 vs 1 \\ \hline
		\textbf{ROBIN(S) vs ROBIN(D)}                                                                                   & 26                                                                                            & 6 vs 0 & 6 vs 0 & 6 vs 0 \\ \hline
		\textbf{K-Means++ vs DK-Means++}                                                                                 & 26                                                                                            & 6 vs 2 & 6 vs 2 & 7 vs 2 \\ \hline
	\end{tabular}
	\caption{\textbf{Comparisons on the maximum performance between stochastic and deterministic methods.} Each plot shows the performance of the Hartigan-Wong's K-Means clustering solution using the purity corresponding to the best silhouette score achieved within 50 different executions (y-axis) on different data sets models (x-axis) and initialized with different stochastic methods (only the cases where significant difference was present are shown). Purity for best silhouette score was averaged over the 40 data sets for each model (gap, weighted gap and Brodinova). The errorbars are showing the standard deviation across the 40 data sets. Solid lines on any two bars underline the level of significant difference between the corresponding methods (only cases with significant difference are showed). The accompanied Table below the figure shows a summary of the comparisons through all the K-Means variations (Hartigan-Wong's K-Means (HW), Lloyd's K-Means (Ll) and K-Medians (KMed)) where there is a significant performance difference between the compared methods.} \label{fig5}  
\end{figure}

Afterwards, we compared the maximum performance of stochastic methods with the performance of deterministic methods similarly to our previous experiment (refer to Figure \ref{fig3}). We compare the stochastic and deterministic variations of Maximin and ROBIN as well as the best stochastic performer of the current experiment (K-Means++) and the best deterministic performer (DK-Means++ see Figure \ref{fig2}). Based on the results in Figure \ref{fig5}, stochastic variations of Maximin and ROBIN achieve overall better performance than their deterministic counterparts and K-Means++ is better than DK-Means++.

We also compare the K-Means variations using different intialisation methods. Based on the result on Table \ref{Talgos} K-Medians achieves the best performance followed by Hartigan-Wong's; Lloyd's was the worst performer. Nevertheless, these systematic differences correspond to only $1.5\%$ purity difference.

%We should highlight that these observed differences, though systematic, are very small, corresponding to $1.5\%$ purity difference.

\begin{table}[h]
	\caption{\textbf{Best performance comparison on K-Means variations using different initialisation methods.} Each row compares two K-Means variations (Hartigan-Wong's K-Means: HW, Lloyd's K-Means: Ll and K-Medians: KMed) initialised with the same method on 26 occasions (10 gap and weighted gap, 12 Brodinova and 4 mixed models). The comparison is based on the times that there was significant difference between the two methods on their maximum performance based on the purity for the best silhouette score. 
		%This score was computed by obtaining over 50 executions the best execution of each stochastic method and matching it to its respective purity and then averaging over the 40 data sets of each model (for deterministic methods this is the average purity over the 40 data sets of each model).
	} \label{Talgos}
	\centering
	\begin{tabular}{|l|c|c|c|c|}
		\hline
		\multicolumn{1}{|c|}{} &
		&
		\multicolumn{3}{c|}{\textbf{\begin{tabular}[c]{@{}c@{}}Significantly better\\ maximum performance\end{tabular}}} \\ \cline{3-5} 
		\multicolumn{1}{|c|}{\multirow{-2}{*}{\textbf{\begin{tabular}[c]{@{}c@{}}Initialization\\ method\end{tabular}}}} &
		\multirow{-2}{*}{\textbf{\begin{tabular}[c]{@{}c@{}}Total number\\ of instances\end{tabular}}} &
		\textbf{HW vs Ll} &
		\textbf{HW vs KMed} &
		\textbf{Ll vs KMed} \\ \hline
		\textbf{Random}     & 26 & 4 vs 1 & 3 vs 7 & 1 vs 7 \\ \hline
		\textbf{K-Means++}  & 26 & 5 vs 1 & 2 vs 7 & 1 vs 4 \\ \hline
		\textbf{ROBIN(S)}      & 26 & 4 vs 0 & 1 vs 3 & 1 vs 7 \\ \hline
		\textbf{Maximin(S)} & 26 & 2 vs 1 & 3 vs 5 & 2 vs 3 \\ \hline
		\rowcolor[HTML]{000000} 
		{\color[HTML]{FFFFFF} \textbf{Total}} &
		{\color[HTML]{FFFFFF} \textbf{104}} &
		{\color[HTML]{FFFFFF} \textbf{15 vs 3}} &
		{\color[HTML]{FFFFFF} \textbf{9 vs 22}} &
		{\color[HTML]{FFFFFF} \textbf{5 vs 21}} \\ \hline
		\textbf{Kaufman}    & 26 & 1 vs 1 & 2 vs 6 & 1 vs 6 \\ \hline
		\textbf{DK-Means++} & 26 & 1 vs 0 & 2 vs 4 & 1 vs 5 \\ \hline
		\textbf{ROBIN(D)}   & 26 & 6 vs 0 & 2 vs 2 & 1 vs 6 \\ \hline
		\textbf{Maximin(D)} & 26 & 7 vs 0 & 2 vs 2 & 1 vs 6 \\ \hline
		\rowcolor[HTML]{000000} 
		{\color[HTML]{FFFFFF} \textbf{Total}} &
		{\color[HTML]{FFFFFF} \textbf{104}} &
		{\color[HTML]{FFFFFF} \textbf{15 vs 1}} &
		{\color[HTML]{FFFFFF} \textbf{8 vs 14}} &
		{\color[HTML]{FFFFFF} \textbf{4 vs 23}} \\ \hline
	\end{tabular}
\end{table}

\subsection{Standalone synthetic and real-world data sets}

We regard the standalone data sets as cases where supervised information is unknown and we assess the performance of the algorithms based on the silhouette index. Detailed results for each data set (minimum, maximum, average performance and variance for each K-Means variation) are illustrated in the supplementary material. Given that these data sets are unique it is difficult to draw definite conclusions similar to the ones where data set models were used but we would like to highlight some observations. DK-Means++ was always able to achieve the best performance of the unsupervised methods while ROBIN(D) failed to achieve the best performance in the cases of A-sets 1, S-Sets 3 and S-Sets 4 when Lloyds and K-Medians were considered; Kaufman and Maximin(D) where the worst performers. From the stochastic methods ROBIN(S) always managed to achieve the maximum performance apart from one case of S-Sets 3 when the Harigan-Wong K-Means was considered; Random was the worst performer. For the real-world data sets most algorithms behaved the same but Maximin(S) outperformed everyone else in the cases of Yeast (all K-Means variations) and Ionosphere (Lloyd's K-Means only). In the case of Glass (all K-Means variations) K-Means++ and Maximin(S) had the best performances. However, with the real data sets we should consider the fact that in rare situations the number of clusters equals to the number of classes \cite{gehring2015detailed} thus it might not be the best examples for clustering benchmarking. Also the relatively better performance of Maximin(S) appears only in these few cases where the Silhouette index indicates poor clustering results in general. In such cases comparative conclusions may not be meaningful as these specific results could be a product of chance.

\subsection{Average number of runs for which stochastic methods reach or surpass deterministic methods}	

In the aforementioned experiments we considered 50 executions of the clustering algorithm using stochastic methods. On average deterministic methods provide better results than stochastic methods but overall stochastic methods may lead to a better clustering solution. We would therefore like to quantify how often this happens. To estimate this, we divide the number of total repetitions (50) by the number of cases where the stochastic method performed better than the deterministic. Table \ref{invprob} summarises the results of this analysis on selected data sets based on their size, dimensionality and number of clusters among two stochastic (Random and K-Means++) and two deterministic methods (DK-Means++ and ROBIN(D)). Based on the results the number of repetitions required for the clustering method using K-Means++ in order to match or surpass the performance of DK-Means++ and ROBIN(D) are less compared with Random. This was expected given the performance comparison of Random and K-Means++ but an important result is the following: there are cases (Yeast, A-sets 2 and A-sets 3) where these stochastic methods fail to match or surpass the performance of deterministic methods under 50 runs. We also observe that when the size of the data set surpasses the 1000 data points the number of required repetitions is significantly high. Finally we should mention that these performance differences are minor, in the order of $10^{-3}$ (purity) on average.

\begin{table}
	\centering
	\begin{tabular}{|l|c|c|c|c|c|}
		\hline
		\multicolumn{1}{|c|}{} & \textbf{\begin{tabular}[c]{@{}c@{}}Rand\\ DKM++\end{tabular}} & \textbf{\begin{tabular}[c]{@{}c@{}}Rand\\ ROBIN(D)\end{tabular}} & \textbf{\begin{tabular}[c]{@{}c@{}}KM++\\ DKM++\end{tabular}} & \textbf{\begin{tabular}[c]{@{}c@{}}KM++\\ ROBIN(D)\end{tabular}} & \textbf{\begin{tabular}[c]{@{}c@{}}size, \\ dimensions,\\ number of \\ clusters\end{tabular}} \\ \hline
		\textbf{gap 2}         & 5                                                                    & 5                                                                  & 4                                                                       & 4                                                                     & 100,2,3                                                                                \\ \hline
		\textbf{wgap 2}        & 6                                                                    & 6                                                                  & 4                                                                       & 4                                                                     & 115,2,2                                                                                \\ \hline
		\textbf{wgap 4}        & 6                                                                    & 6                                                                  & 5                                                                       & 5                                                                     & 200,2,2                                                                                \\ \hline
		\textbf{wgap 3}        & 3                                                                    & 3                                                                  & 3                                                                       & 3                                                                     & 200,2,4                                                                                \\ \hline
		\textbf{gap 5}         & 4                                                                    & 2                                                                  & 4                                                                       & 2                                                                     & 200,3,2                                                                                \\ \hline
		\textbf{wgap 6}        & 8                                                                    & 8                                                                  & 7                                                                       & 6                                                                     & 300,2,6                                                                                \\ \hline
		\textbf{gap 3}         & 5                                                                    & 5                                                                  & 4                                                                       & 4                                                                     & \textit{143,3,4}                                                                       \\ \hline
		\textbf{gap 4}         & 10                                                          & 10                                                       & 6                                                                       & 6                                                                     & \textit{158,10,2}                                                                      \\ \hline
		\textbf{wgap 1}        & 7                                                                    & 7                                                                  & 6                                                                       & 6                                                                     & \textit{227,2,6}                                                                       \\ \hline
		\textbf{wgap 5}        & \textbf{14}                                                          & \textbf{14}                                                        & 6                                                                       & 6                                                                     & \textit{141,10,2}                                                                      \\ \hline
		\textbf{brod 1}        & 4                                                                    & 4                                                                  & 5                                                                       & 5                                                                     & 120,20,3                                                                               \\ \hline
		\textbf{brod 2}        & \textbf{18}                                                          & \textbf{17}                                                        & \textbf{17}                                                             & \textbf{17}                                                           & 400,20,3                                                                               \\ \hline
		\textbf{Iris}          & 7                                                                    & 7                                                                  & 3                                                                       & 3                                                                     & 150,4,3                                                                                \\ \hline
		\textbf{Wine}          & 3                                                                    & 3                                                                  & 2                                                                       & 2                                                                     & 178,13,3                                                                               \\ \hline
		\textbf{Glass}         & 7                                                                    & 7                                                                  & 6                                                                       & 6                                                                     & 214,9,6                                                                                \\ \hline
		\textbf{Ionosphere}    & 3                                                                    & 3                                                                  & 2                                                                       & 2                                                                     & 351,34,2                                                                               \\ \hline
		\textbf{\begin{tabular}[c]{@{}l@{}}Breast\\ Cancer\end{tabular}}  & \textbf{12}                                                          & \textbf{12}                                                        & 7                                                                       & 7                                                                     & 683,9,2                                                                                \\ \hline
		\textbf{Yeast}         &  \textbf{N/A}                                                                 & \textbf{16}                                                        & \textbf{27}                                                             & \textbf{14}                                                           & 1484,8,10                                                                              \\ \hline
		\textbf{A-sets 1}      & \textbf{28}                                                          & \textbf{26}                                                        & \textbf{26}                                                             & \textbf{16}                                                           & 3000,2,15                                                                              \\ \hline
		\textbf{S-Sets 1}      & \textbf{34}                                                          & \textbf{34}                                                        & \textbf{26}                                                             & \textbf{15}                                                           & 5000,2,15                                                                              \\ \hline
		\textbf{A-sets 2}      & \textbf{N/A}                                                                  &  \textbf{N/A}                                                               & \textbf{34}                                                             & \textbf{33}                                                           & 5250,2,35                                                                              \\ \hline
		\textbf{A-sets 3}      & \textbf{N/A}                                                                  & \textbf{N/A}                                                                & \textbf{N/A}                                                                     & \textbf{N/A}                                                                   & 7500,2,50                                                                              \\ \hline
	\end{tabular}
	\caption{\textbf{Average number of runs for which stochastic initialisations achieve equivalent or better performance than deterministic initialisations.} Each column shows a comparison between clustering initialised with stochastic and deterministic methods (Rand = Random, DKM++ = DK-Means++, KM++ = K-Means++). Each cell value corresponds to the number of executions of the K-Means clustering initialised with the stochastic method to reach or surpass its performance if it was initialised with the deterministic method and executed once. N/A values mean that in these occasions the stochastic clustering was not able to match or surpass the performance of the deterministic clustering under $50$ executions. Values higher than $10$ are marked in bold. The data sets are arranged based on their \textit{size}, \textit{dimensionality} and \textit{number of clusters} (see info on last column; numbers in italics correspond to the average number of elements should the model generated data sets of different sizes.} \label{invprob}  
\end{table}

\subsection{Execution time analysis}

Finally, we performed an execution time analysis on the initialisation methods using a selection of the data sets depending on their size, dimensionality and number of clusters; data sets with equivalent properties were omitted. The analysis was performed as follows: each initialisation methods followed by K-Means clustering (Lloyd's K-Means) was executed 50 times and the average running time was taken into consideration. %Specifically for the stochastic processes with linear complexity (Random, K-Means++, Maximin(S)) each average was repeated 50 times and we show estimations of execution time (refer to Figure \ref{time1}) for both 1 and 10 runs. We find that after these number of runs stochastic methods tend to have approximately the same execution time requirements as deterministic methods. For this analysis we should consider the following aspects: 
\begin{itemize}
	\item[$\bullet$] The benchmarking was exclusively performed on a \textit{personal laptop} with the following properties: Dell G7; Intel i7-9750H processor; 16 GB RAM; Windows 10 Pro edition.
	\item[$\bullet$] All the algorithms were written in MATLAB but the LOF score for ROBIN was computed using R code (specifically the dbscan package \cite{Rcode}) because we found that the MATLAB implementation was very slow.
	\item[$\bullet$] The running time recording includes the initialisation method and the clustering algorithm. For ROBIN the computation of LOF was included in the execution duration as well as the computation of $\varepsilon$ for the DK-Means++.
\end{itemize}

\begin{figure}[h]
	\centering
	\includegraphics[width=\linewidth]{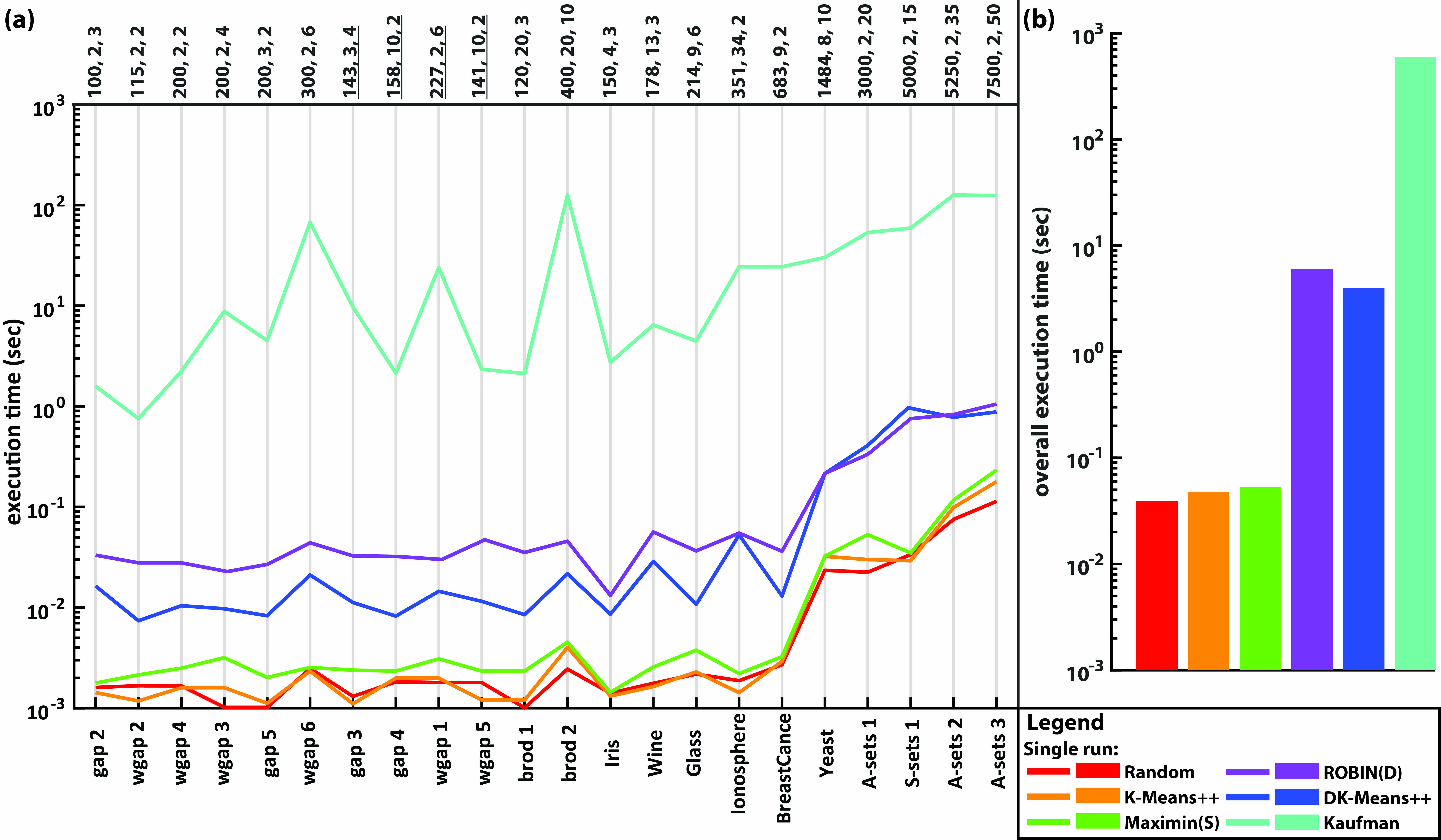}
	\caption{\textbf{Execution time analysis.} \textbf{(a)} Each line shows the execution duration of an initialisation method on different data sets selected based on size, dimensions and number of clusters. We average the execution time for 50 repetitions and, in case of models, across the 40 data sets of each model. The data sets are arranged based on their \textit{size}, \textit{dimensionality} and \textit{number of clusters} (see info on top, underlined numbers means that for these models generate data sets of different sizes). \textbf{(b)} Each bar shows the summed execution time across all data sets of (a).} \label{time1}  
\end{figure}

\noindent	Based on the results in Figure \ref{time1} Kaufman is the worst method in terms of execution duration and it is affected both by the size, dimensionality and number of clusters. Random and K-Means++ are the fastest methods followed by Maximin(S). DK-Means++ is almost always better than ROBIN(D) in terms of speed for our implementation. %When considering 10 executions of the stochastic methods (Random, K-Means++ and Maximin(S)) we observed that ROBIN(D) and specifically DK-Means++ are not that much slower than Random, K-Means++ and Maximin(S) and sometimes surpass them in speed. 

\begin{figure}[h]
	\centering
	\includegraphics[width=\linewidth]{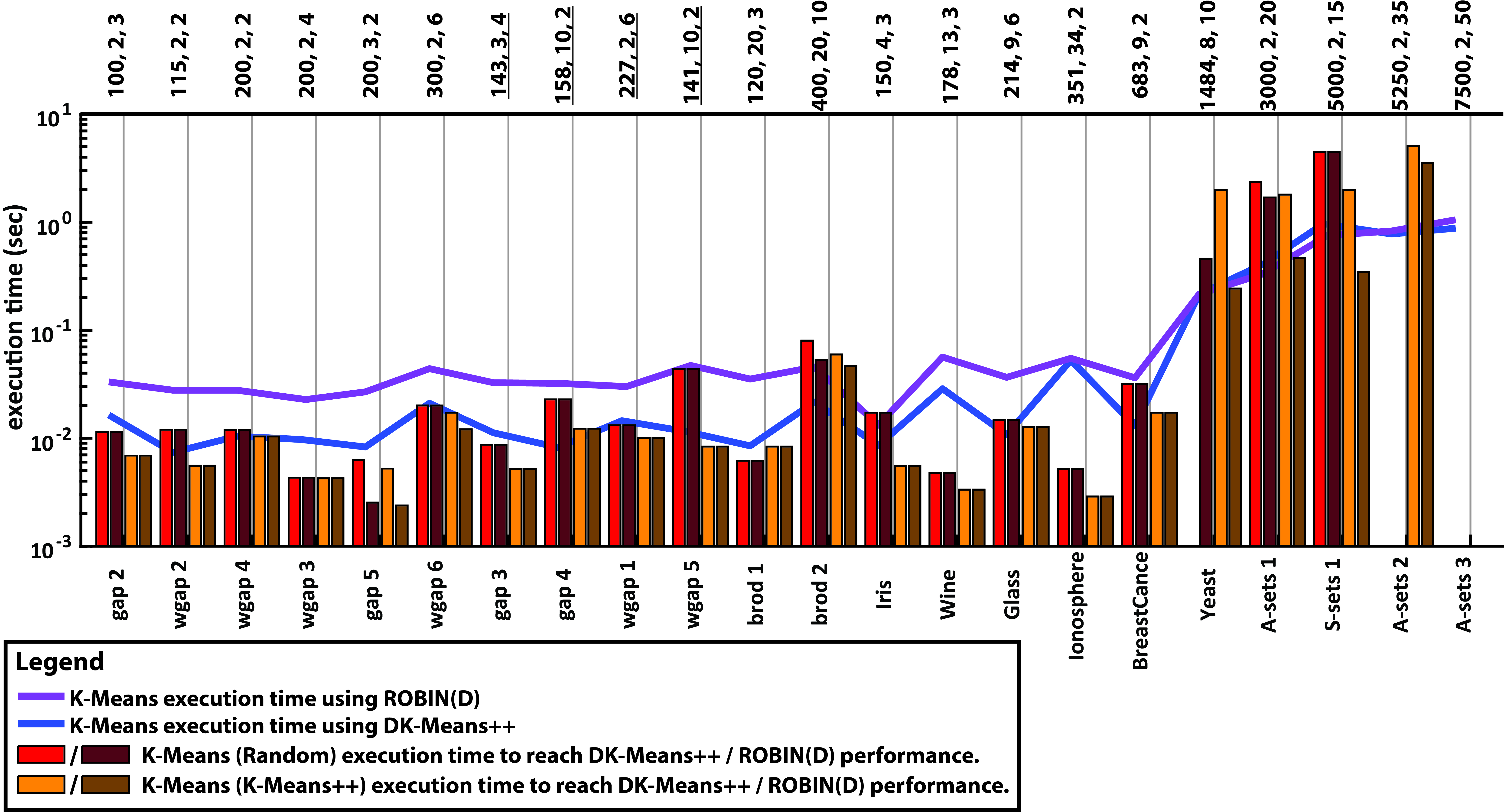}
	\caption{\textbf{Execution time until stochastic methods reach or surpass the performance of deterministic methods.} Each bar shows the running time requirements of the clustering algorithm initialised with a stochastic method to reach or surpass the performance of the  same algorithm initialised with a deterministic method. The time requirements of the clustering algorithm using a deterministic method are shown as lines for comparison. Cases where bars are not shown mean that, up to 50 runs, the clustering algorithm using a stochastic method was unable to surpass the performance of the deterministic method.} \label{time2}  
\end{figure}

Furthermore, and based on the results of Table \ref{invprob} we performed an analysis on the time requirements of the stochastic methods to reach or surpass the performance of deterministic methods with multiple executions of the clustering algorithm using different seeds. In Figure \ref{time2} we show the single run execution time of the stochastic initialisation (plus the clustering overhead) multiplied by the number of iterations required to surpass the deterministic methods (see Table \ref{invprob}). Based on the results shown in Figure \ref{time2} we observe that in many occasions running DK-Means++ once is better in terms of execution time than repeated runs of the clustering with a stochastic method. Equivalent conclusions can also be obtained from the Maximin(S) initialisation method (supplementary material).

\noindent We also investigated the number of iterations required for the clustering algorithm to reach convergence using stochastic and deterministic methods. We find that stochastic methods overall provide worst initial conditions (with the exception of ROBIN(S) vs Maximin(D)), and as a consequence  the clustering algorithm requires more iterations to converge, which adds to the overhead of the stochastic initialisation methods (for results refer to the supplementary material).

\section{Discussion}
K-Means clustering remains one of the most common clustering techniques in many different research fields and frequently it is used as a component of more complex algorithms (e.g. hierarchical clustering \cite{jain2010data}). Following similar benchmark studies on K-Means \cite{celebi2013comparative,franti2018k,franti2019much}, in this study we compare stochastic and deterministic initialisation methods on K-Means variations. We particularly investigated the methods of ROBIN and DK-Means++ since to the best of our knowledge they have not been studied as extensively as other initialisation methods. Experimentally we showed:
\begin{itemize}
	\item[$\bullet$] More sophisticated initialisation methods can lead, on average, to better clustering regardless of the K-Means variation (see Table \ref{TalgosMean}). From the stochastic methods, ROBIN(S) can achieve the best average performance compared with Random, K-Means++ and Maximin(S) (see Figure \ref{fig1}). From the deterministic methods, DK-Means++ can achieve the best performance compared with Kaufman, Maximin(D) and ROBIN(D) (see Figure \ref{fig2}). In addition, DK-Means++ can achieve better performance from the average performance of stochastic methods (see Figure \ref{fig3}). Overall, deterministic methods have on average less performance variability across the data sets of each model we tested and lead to more stable solutions than stochastic methods (see supplementary material) and can surpass the performance of stochastic methods (see Figure \ref{fig3}).
	
	\item[$\bullet$] When executed multiple times stochastic methods can achieve better performance than deterministic methods. Opposite the the first point, in that case, less sophisticated methods (such as Random and K-Means++ as opposed to ROBIN(S)) can achieve better performance (see Figure \ref{fig5}). K-Means++ with 50 executions achieved the best performance followed by Random (see Figure \ref{fig4}). The only deterministic method that can still compete to an extent is DK-Means++ (see supplementary material where we provide a full list of all comparisons among all the initialisation methods). 
	
	\item[$\bullet$] 
	We found (see Table \ref{Talgos}) that as indicated by \cite{slonim2013hartigan} Hartigan-Wong K-Means is better than Lloyd's K-Means and as shown by \cite{brusco2017comparison} (only for one K-Means variant) K-Medians is better than both Hartigan-Wong and Lloyd's K-Means. However these differences add up to performance difference of only $1.5\%$ as measured by the purity index.
	
	\item[$\bullet$] Regarding execution time requirements, Random and K-Means++ are fastest performers in terms of single runs while Kaufman the slowest (see Figure \ref{time1}). Maximin(S) is slightly slower than K-Means++. Nevertheless these methods require multiple executions in order to reach the performance of determinsitic methods (refer to Table \ref{invprob}) especially with bigger data sets (number of elements to thousands). Multiple executions of these methods have almost similar requirements as a single run of deterministic methods DK-Means++ and ROBIN(D) (refer to Figure \ref{time2}). This is due to the fact that the clustering algorithm requires more iterations to reach converge when stochastic methods are used (refer to the supplementary material). Between DK-Means++ and ROBIN(D) the former is faster than the latter. 
\end{itemize}

Overall, and from a practical point of view, the stochastic Random and the deterministic Kaufman methods are not advisable. The first method despite being the simplest and the fastest can be replaced with K-Means++ that has better probability of achieving superior performance. The latter method is extremely slow and there are better alternatives such as the DK-Means++ that has both better performance and execution time. Maximin(D) and ROBIN(S) are not advisable either since the former is relatively fast and multiple executions of Maximin(S) can be performed instead while the latter has much more time requirements, small variability on its solutions and when an approximate clustering is required ROBIN(D) can be used instead. DK-Means++ is a good option when determinism is required since with a single run it can achieve better performance compared with other dete trministic methods and comparable performance to multiple executions of stochastic methods that would require the same or more running time. In applications where exhausted search of optimal initial centroids needs to be performed K-Means++ should be considered (the study of \cite{celebi2013comparative} has also benchmarked a greedy version of this method which is also recommended). In these cases, if time requirements are flexible, a strategy would be to perform first DK-Means++ which would give an indication about the clustering capabilities of the data set and then multiple executions of K-Means++. We should add that in the mixed model 4, ROBIN(S) and ROBIN(D) performed significantly low compared with other cases because both where  placing two initial centroids on the sides of the most elongated cluster while DK-Means++ were placing correctly a centroid almost in the middle of the cluster. This indicates that the DK-Means++'s heuristic might be more robust to applications than the LOF score of ROBIN for clusters detection. It should be noted that more complex techniques like DK-Means++ can be considered as clustering algorithms themselves since they produce good initial clusters. This observation was mentioned in the study of \cite{celebi2013comparative} and we also performed a study (refer to the supplementary material) on the number of iterations until the K-Means algorithm reaches convergence when it is initialized with different methods. Based on the results deterministic methods cause K-Means to converge faster than when it is initialized with stochastic methods. As expected, ROBIN(S) and DK-Means++ were again the best stochastic and deterministic methods on that account.

These conclusions were based on extensive benchmarking considering many different clustering models from other studies: Gaussian, high-dimensional (10 dimensions), elongated, unbalanced, non-Gaussian from the studies of \cite{tibshirani2001estimating} and \cite{yan2007determining}; high-dimensional (20 dimensions) containing informative and uninformative features and higher-dimensional (1000 and 1500 dimensions) with varying number of clusters (3, 10, 50 clusters) and cluster sizes (50-100 points) \cite{brodinova2017robust}. We also used our own models which contain clusters with different properties (unbalanced, elongated and Gaussian; unbalanced Gaussian and non-Gaussian; unbalanced, Gaussian with different variability among their dimensions). 

With the use of synthetic data set generators we had the ability to generate multiple data sets and run hypothesis testing to further strengthen our conclusions but we also considered standalone data sets. The ``clustering data sets'' S-sets \cite{Ssets} and A-sets \cite{Asets} were selected from the studies of \cite{franti2018k,franti2019much} because both are containing more clusters and data points than the generated ones and also because in the case of the S-sets the clusters are having different overlap degrees. The conclusions we obtained from the data set generators match with the conclusions of the standalone S-sets and A-sets data sets. Specifically for our higher dimensional data sets (1000, 1500 dimensions) generated using the Brodinova generator \cite{brodinova2017robust} (see Table \ref{sdatasets3}) we selected to have small clusters due to the Kaufman initialization method which requires significant amount of time to be executed. However, we also generated data sets with larger clusters (approximately five times bigger) and we tested the ROBIN(D) and DK-Means++ methods on them. The results (not shown) and conclusions were similar to the ones reported already.

Based on the previous studies \cite{franti2018k,franti2019much} the authors have clearly demonstrated that K-Means performs worse when there is large number of clusters and that dimensionality does not have a direct effect on the performance of the algorithm. In our experiments using the Brodinova models (see Figures \ref{fig2}, \ref{fig5} brod 1 to brod 12) we observe that indeed the performance of all the methods drops when the number of clusters is increased regardless of the dimensionality, especially in the case of Brodinova brod10 model where we generate data sets having 50 clusters. Apart from the last extreme case, we observed that multiple executions of stochastic methods improve the performance of K-Means. It should also be noted that the deterministic DK-Means++ method achieves (similar to multiple executions of stochastic methods i.e. Random, K-Means++, Maximin(S) and ROBIN(S)) the highest performance on the clustering basic benchmark \cite{franti2018k,franti2019much} in all the cases (see the supplementary material) even though these data sets have high number of clusters (A-sets: 20, 30, 50; S-sets: 15). The same authors \cite{franti2018k,franti2019much} also demonstrated that strong cluster unbalances affect negatively the K-Means clustering. In our experiments and specifically for the weighted gap 2 model we observed that data sets with unbalanced clusters do not cause any particular issues to the maximum performances of the algorithms. For the performance between K-Means and K-Medians, similar to the results of \cite{brusco2017comparison}, we found that K-Medians outperforms K-Means on synthetic data set models but on a small difference of $1\%$ of purity and on standalone data sets (both synthetic and real-world) any particular differences among the K-Means variations couldn't be clearly detected.

In order to show application to ``real world problems'' previous studies have chosen to use standard classification data sets as benchmarks for clustering. While this approach is commonly used, in these data the mapping from classes to clusters is somehow forced: it is possible that data from one class belong to different clusters, and assuming that number of clusters equals number of classes is likely to underestimate the true number of clusters. This can be seen from the low value of the Silhouette index especially in the cases of Ionosphere and Yeast data sets. For this we base our conclusions mostly on the benchmark models that allows us to generate multiple samples and evaluate the statistical significance of the results. In fact, we considered a broad combination of different clusters, in terms of normality (Gaussian, non-Gaussian), shape (spherical, elongated) and size (clusters with different number of data points) including high dimensional data, as found in real world applications such as bioinformatics \cite{wang2008approaches}. 

It should also be noted that many clustering frameworks designed to deal with complex data sets (e.g. sub-clustering \cite{biswas2014active}, or sparse clustering \cite{witten2010framework,kondo2016rskc,brodinova2017robust}) are using the K-Means or some variant of it and are dependent on good clustering initialisation. Our experimental work revealed that there are deterministic methods (DK-Means++ \cite{nidheesh2017enhanced}) that lead to a good clustering solution with a single execution of the K-Means algorithm.

A limitation of the current study is that the execution time analysis is subject to the machine that executed it. More powerful machines or code optimisation of the algorithms and initialisation methods can change time analysis results. Nevertheless the rest of the analysis including the number of different seeds for stochastic methods to reach the performance of deterministic is, on average, reproducible. Statistics on average performance comparison are representative since, the analysis of sections 4.1, 4.2 and 4.3 had been also performed on 25 instances of the various data sets models instead of 50 and led to the same conclusions.

\section*{Contributions}
A.V. performed the experiments and wrote the manuscript. E.V. contributed to the statistical and algorithmic analysis, provided feedback and edited the manuscript. S.L. oversaw the benchmarking process, provided feedback on the results presentation and the manuscript. M.C. contributed to the literature review and the discussion and performed the final proofreading.

\section*{Acknowledgments}
This research was funded by the Numerical Algorithms Group (NAG). We would also like to thank the reviewers for their detailed and constructive comments especially on the execution time considerations and on the inclusion of both a supervised and unsupervised benchmarking comparison.

\bibliographystyle{unsrt}  
\bibliography{references}

\cleardoublepage
\pagenumbering{arabic}
\setcounter{page}{1}
\setcounter{figure}{0}   
\setcounter{table}{0}   
\appendix

\begin{table}
	\centering
	\caption{\textbf{Summary of comparisons on average performance of stochastic and deterministic methods over different K-Means variations on synthetic data set models.} In the first part of the table, each row compares two different methods over the Hartigan-Wong's K-Means (HW), Lloyd's K-Means (Ll) and K-Medians (KMed) algorithms on 26 occasions (10 gap and weighted gap, 12 Brodinova and 4 mixed models). The comparison is separate among the stochastic and deterministic methods and based on the times that there was significant difference between the two methods over the 40 data sets of each model. Based on the results ROBIN(S) is the best performer of stochastic methods and DK-Means the best performer of deterministic methods, both over all the clustering algorithms. The second part of the table groups all the stochastic and deterministic methods together and counts the overall percentage of observed significant differences. Based on the results the performance differences among the deterministic methods are less compared to the stochastic methods suggesting less performance variability. This Table accompanies Figure 2 and Figure 3 of the main manuscript which show the average performance of the initialisation methods based on Silhouette and show comparisons only for the Hartigan-Wong's algorithm.}
	\label{exp1T1}
	\begin{tabular}{|l|c|c|c|c|}
		\hline
		\multicolumn{1}{|c|}{}                                                                                                              &                                                                                                & \multicolumn{3}{c|}{\textbf{\begin{tabular}[c]{@{}c@{}}Significant better\\ average performance\\ Silhouette (Purity)\end{tabular}}}                                                    \\ \cline{3-5} 
		\multicolumn{1}{|c|}{\multirow{-2}{*}{\textbf{\begin{tabular}[c]{@{}c@{}}Initialization\\ method\end{tabular}}}}                    & \multirow{-2}{*}{\textbf{\begin{tabular}[c]{@{}c@{}}Total number\\ of instances\end{tabular}}} & \textbf{HW}                                                 & \textbf{Ll}                                                 & \textbf{KMed}                                               \\ \hline
		\textbf{Random vs K-Means++}                                                                                                        & 26                                                                                             & \begin{tabular}[c]{@{}c@{}}0 vs 23\\ (0 vs 21)\end{tabular} & \begin{tabular}[c]{@{}c@{}}0 vs 23\\ (0 vs 22)\end{tabular} & \begin{tabular}[c]{@{}c@{}}0 vs 23\\ (0 vs 24)\end{tabular} \\ \hline
		\textbf{Random vs ROBIN(S)}                                                                                                         & 26                                                                                             & \begin{tabular}[c]{@{}c@{}}1 vs 22\\ (1 vs 22)\end{tabular} & \begin{tabular}[c]{@{}c@{}}3 vs 22\\ (3 vs 22)\end{tabular} & \begin{tabular}[c]{@{}c@{}}2 vs 23\\ (2 vs 23)\end{tabular} \\ \hline
		\textbf{Random vs Maximin(S)}                                                                                                       & 26                                                                                             & \begin{tabular}[c]{@{}c@{}}0 vs 17\\ (6 vs 15)\end{tabular} & \begin{tabular}[c]{@{}c@{}}1 vs 19\\ (6 vs 16)\end{tabular} & \begin{tabular}[c]{@{}c@{}}1 vs 19\\ (6 vs 16)\end{tabular} \\ \hline
		\textbf{K-Means++ vs ROBIN(S)}                                                                                                      & 26                                                                                             & \begin{tabular}[c]{@{}c@{}}1 vs 22\\ (1 vs 21)\end{tabular} & \begin{tabular}[c]{@{}c@{}}3 vs 22\\ (3 vs 21)\end{tabular} & \begin{tabular}[c]{@{}c@{}}2 vs 23\\ (2 vs 22)\end{tabular} \\ \hline
		\textbf{K-Means++ vs Maximin(S)}                                                                                                    & 26                                                                                             & \begin{tabular}[c]{@{}c@{}}5 vs 14\\ (8 vs 12)\end{tabular} & \begin{tabular}[c]{@{}c@{}}5 vs 16\\ (9 vs 14)\end{tabular} & \begin{tabular}[c]{@{}c@{}}5 vs 15\\ (7 vs 13)\end{tabular} \\ \hline
		\textbf{ROBIN(S) vs Maximin(S)}                                                                                                     & 26                                                                                             & \begin{tabular}[c]{@{}c@{}}17 vs 1\\ (17 vs 1)\end{tabular} & \begin{tabular}[c]{@{}c@{}}17 vs 3\\ (18 vs 3)\end{tabular} & \begin{tabular}[c]{@{}c@{}}18 vs 1\\ (19 vs 1)\end{tabular} \\ \hline
		\rowcolor[HTML]{000000} 
		& \multicolumn{1}{l|}{\cellcolor[HTML]{000000}}                                                  & \multicolumn{1}{l|}{\cellcolor[HTML]{000000}}               & \multicolumn{1}{l|}{\cellcolor[HTML]{000000}}               & \multicolumn{1}{l|}{\cellcolor[HTML]{000000}}               \\ \hline
		\textbf{Kaufman vs DK-Means++}                                                                                                      & 26                                                                                             & \begin{tabular}[c]{@{}c@{}}3 vs 8\\ (3 vs 9)\end{tabular}   & \begin{tabular}[c]{@{}c@{}}3 vs 9\\ (3 vs 10)\end{tabular}  & \begin{tabular}[c]{@{}c@{}}3 vs 7\\ (4 vs 8)\end{tabular}   \\ \hline
		\textbf{Kaufman vs ROBIN(D)}                                                                                                        & 26                                                                                             & \begin{tabular}[c]{@{}c@{}}5 vs 8\\ (4 vs 8)\end{tabular}   & \begin{tabular}[c]{@{}c@{}}6 vs 8\\ (6 vs 6)\end{tabular}   & \begin{tabular}[c]{@{}c@{}}5 vs 6\\ (4 vs 6)\end{tabular}   \\ \hline
		\textbf{Kaufman vs Maximin(D)}                                                                                                      & 26                                                                                             & \begin{tabular}[c]{@{}c@{}}8 vs 5\\ (8 vs 5)\end{tabular}   & \begin{tabular}[c]{@{}c@{}}10 vs 6\\ (10 vs 6)\end{tabular} & \begin{tabular}[c]{@{}c@{}}9 vs 5\\ (9 vs 5)\end{tabular}   \\ \hline
		\textbf{DK-Means++ vs ROBIN(D)}                                                                                                     & 26                                                                                             & \begin{tabular}[c]{@{}c@{}}2 vs 1\\ (4 vs 0)\end{tabular}   & \begin{tabular}[c]{@{}c@{}}5 vs 1\\ (6 vs 0)\end{tabular}   & \begin{tabular}[c]{@{}c@{}}4 vs 1\\ (6 vs 1)\end{tabular}   \\ \hline
		\textbf{DK-Means++ vs Maximin(D)}                                                                                                   & 26                                                                                             & \begin{tabular}[c]{@{}c@{}}7 vs 0\\ (9 vs 0)\end{tabular}   & \begin{tabular}[c]{@{}c@{}}9 vs 0\\ (11 vs 0)\end{tabular}  & \begin{tabular}[c]{@{}c@{}}7 vs 0\\ (10 vs 1)\end{tabular}  \\ \hline
		\textbf{ROBIN(D) vs Maximin(D)}                                                                                                     & 26                                                                                             & \begin{tabular}[c]{@{}c@{}}8 vs 1\\ (8 vs 1)\end{tabular}   & \begin{tabular}[c]{@{}c@{}}9 vs 1\\ (10 vs 1)\end{tabular}  & \begin{tabular}[c]{@{}c@{}}8 vs 1\\ (9 vs 1)\end{tabular}   \\ \hline
		\multicolumn{1}{|c|}{}                                                                                                              &                                                                                                & \multicolumn{3}{c|}{\textbf{\begin{tabular}[c]{@{}c@{}}Observed significant\\ performance differences\\ on average performance\\ Silhouette (Purity)\end{tabular}}}                                              \\ \cline{3-5} 
		\multicolumn{1}{|c|}{\multirow{-2}{*}{\textbf{\begin{tabular}[c]{@{}c@{}}Initialization\\ methods\end{tabular}}}}                   & \multirow{-2}{*}{\textbf{\begin{tabular}[c]{@{}c@{}}Total number\\ of instances\end{tabular}}} & \textbf{HW}                                                 & \textbf{Ll}                                                 & \textbf{KMed}                                               \\ \hline
		\multicolumn{1}{|c|}{\textbf{\begin{tabular}[c]{@{}c@{}}Stochastic:\\ Random, K-Means++,\\ ROBIN(S), Maximin(S)\end{tabular}}}      & 156                                                                                            & \begin{tabular}[c]{@{}c@{}}78.8\%\\ (80.1\%)\end{tabular}   & \begin{tabular}[c]{@{}c@{}}86.0\%\\ (87.8\%)\end{tabular}   & \begin{tabular}[c]{@{}c@{}}84.6\%\\ (86.5\%)\end{tabular}   \\ \hline
		\multicolumn{1}{|c|}{\textbf{\begin{tabular}[c]{@{}c@{}}Deterministic:\\ Kaufman, DK-Means++,\\ ROBIN(D), Maximin(D)\end{tabular}}} & 156                                                                                            & \begin{tabular}[c]{@{}c@{}}36.0\%\\ (37.8\%)\end{tabular}   & \begin{tabular}[c]{@{}c@{}}42.9\%\\ (44.2\%)\end{tabular}   & \begin{tabular}[c]{@{}c@{}}36.0\%\\ (41.0\%)\end{tabular}   \\ \hline
	\end{tabular}
\end{table}

\begin{table}
	\caption{\textbf{Detailed comparison on the maximum performance of stochastic methods with the performance of deterministic methods.} Each row compares a stochastic with a deterministic method over the Hartigan-Wong's K-Means (HW), Lloyd's K-Means (Ll) and K-Medians (KMed) algorithms on 26 occasions (10 gap and weighted gap, 12 Brodinova and 4 mixed models). The comparison is based on the times that there was significant difference between the two methods on their maximum performance. Based on the results stochastic methods are better that deterministic on achieving the best performance but the more sophisticated the stochastic method is the less performance difference it achieves compared with deterministc methods for which the opposite is observed. This Table accompanies Figure 5 and Figure 6 of the main manuscript which show the maximum performance of the initialisation methods and show comparisons only for the Hartigan-Wong's algorithm.} \label{exp1T5}
	\centering
	\begin{tabular}{|l|c|c|c|c|}
		\hline
		\multicolumn{1}{|c|}{\multirow{2}{*}{\textbf{\begin{tabular}[c]{@{}c@{}}Initialization\\ method\end{tabular}}}} & \multirow{2}{*}{\textbf{\begin{tabular}[c]{@{}c@{}}Total number\\ of instances\end{tabular}}} & \multicolumn{3}{c|}{\textbf{\begin{tabular}[c]{@{}c@{}}Significant better\\ maximum performance\\ Purity for best Silhouette\end{tabular}}} \\ \cline{3-5} 
		\multicolumn{1}{|c|}{}                                                                                          &                                                                                               & \textbf{HW}                        & \textbf{Ll}                        & \textbf{KMed}                        \\ \hline
		\textbf{Random vs Kaufman}                                                                                      & 26                                                                                            & 9 vs 3                             & 9 vs 4                             & 9 vs 4                               \\ \hline
		\textbf{Random vs DK-Means++}                                                                                   & 26                                                                                            & 5 vs 3                             & 4 vs 2                             & 5 vs 3                               \\ \hline
		\textbf{Random vs ROBIN(D)}                                                                                     & 26                                                                                            & 6 vs 3                             & 8 vs 2                             & 7 vs 2                               \\ \hline
		\textbf{Random vs Maximin(D)}                                                                                   & 26                                                                                            & 10 vs 2                             & 12 vs 1                            & 11 vs 1                              \\ \hline
		\textbf{K-Means++ vs Kaufman}                                                                                   & 26                                                                                            & 10 vs 0                            & 12 vs 0                            & 10 vs 0                              \\ \hline
		\textbf{K-Means++ vs DK-Means++}                                                                                & 26                                                                                            & 6 vs 2                             & 6 vs 2                             & 7 vs 2                               \\ \hline
		\textbf{K-Means++ vs ROBIN(D)}                                                                                  & 26                                                                                            & 8 vs 2                             & 9 vs 2                             & 8 vs 2                               \\ \hline
		\textbf{K-Means++ vs Maximin(D)}                                                                                & 26                                                                                            & 10 vs 1                             & 12 vs 1                            & 11 vs 1                              \\ \hline
		\textbf{ROBIN(S) vs Kaufman}                                                                                    & 26                                                                                            & 9 vs 4                             & 8 vs 6                             & 8 vs 4                               \\ \hline
		\textbf{ROBIN(S) vs DK-Means++}                                                                                 & 26                                                                                            & 0 vs 2                             & 1 vs 3                             & 2 vs 2                               \\ \hline
		\textbf{ROBIN(S) vs ROBIN(D)}                                                                                   & 26                                                                                            & 6 vs 0                             & 6 vs 0                             & 6 vs 0                               \\ \hline
		\textbf{ROBIN(S) vs Maximin(D)}                                                                                 & 26                                                                                            & 8 vs 0                             & 9 vs 1                             & 9 vs 0                               \\ \hline
		\textbf{Maximin(S) vs Kaufman}                                                                                  & 26                                                                                            & 9 vs 4                             & 9 vs 4                            & 8 vs 4                               \\ \hline
		\textbf{Maximin(S) vs DK-Means++}                                                                               & 26                                                                                            & 2 vs 4                             & 2 vs 5                             & 3 vs 5                               \\ \hline
		\textbf{Maximin(S) vs ROBIN(D)}                                                                                 & 26                                                                                            & 4 vs 4                             & 6 vs 4                             & 4 vs 4                               \\ \hline
		\textbf{Maximin(S) vs Maximin(D)}                                                                               & 26                                                                                            & 9 vs 1                             & 11 vs 1                            & 10 vs 1                              \\ \hline
	\end{tabular}
\end{table}

\begin{table}[h]	\centering \caption[ ]{\textbf{Comparison of the initialisation methods on standalone clustering data sets based on Silhouette index.} Each stochastic method (Random, K-Means++, ROBIN(S) and Maximin (S)) was executed 50 times and the minimum, maximum and mean performance is shown followed by the performance variation for three K-Means variations. For the maximum performance the cases where a method has achieved the maximum performance is shown in bold.} \label{Clustering} \resizebox{\columnwidth}{!}{	\begin{tabu}{ll| [3pt]c|c|c|c| [3pt]c|c|c|c| [3pt]c|c|c|c| [3pt]} \tabucline [3pt]{3-14} \cline{3-14}& & \multicolumn{4}{c|[3pt]}{K-Means (Hartigan-Wong)} & \multicolumn{4}{c|[3pt]}{K-Means (Lloyd)} & \multicolumn{4}{c|[3pt]}{K-Medians} \\ \cline{3-14}& & min & max & mean & std & min & max & mean & std & min & max & mean & std\\ \tabucline [3pt]{-}\multicolumn{1}{|[3pt]l|}{\multirow{8}{*}{\rotatebox[origin=c]{90}{A-sets 1}}} &Random & 0.446 & \textbf{0.595} & 0.521 & 0.029 & 0.475 & \textbf{0.595} & 0.518 & 0.031 & 0.434 & 0.57 & 0.507 & 0.036 \\ \cline{2-14}\multicolumn{1}{|[3pt]l|}{} & K-Means++ & 0.482 & \textbf{0.595} & 0.548 & 0.03 & 0.484 & \textbf{0.595} & 0.548 & 0.024 & 0.469 & \textbf{0.595} & 0.531 & 0.029 \\ \cline{2-14}\multicolumn{1}{|[3pt]l|}{} & ROBIN(S) & 0.567 & \textbf{0.595} & 0.586 & 0.013 & 0.568 & \textbf{0.595} & 0.58 & 0.014 & 0.567 & \textbf{0.595} & 0.585 & 0.013 \\ \cline{2-14}\multicolumn{1}{|[3pt]l|}{} & Maximin(S) & 0.52 & \textbf{0.595} & 0.56 & 0.019 & 0.519 & \textbf{0.595} & 0.56 & 0.023 & 0.499 & \textbf{0.595} & 0.553 & 0.03 \\ \cline{2-14}\multicolumn{1}{|[3pt]l|}{} & Kaufman & 0.567 & 0.567 & 0.567 & 0 & 0.567 & 0.567 & 0.567 & 0 & 0.567 & 0.567 & 0.567 & 0 \\ \cline{2-14}\multicolumn{1}{|[3pt]l|}{} & DK-Means++ & 0.595 & \textbf{0.595} & 0.595 & 0 & 0.595 & \textbf{0.595} & 0.595 & 0 & 0.595 & \textbf{0.595} & 0.595 & 0 \\ \cline{2-14}\multicolumn{1}{|[3pt]l|}{} & ROBIN(D) & 0.567 & 0.567 & 0.567 & 0 & 0.568 & 0.568 & 0.568 & 0 & 0.567 & 0.567 & 0.567 & 0 \\ \cline{2-14}\multicolumn{1}{|[3pt]l|}{} & Maximin(D) & 0.556 & 0.556 & 0.556 & 0 & 0.556 & 0.556 & 0.556 & 0 & 0.538 & 0.538 & 0.538 & 0\\ \tabucline [3pt]{-}\multicolumn{1}{|[3pt]l|}{\multirow{8}{*}{\rotatebox[origin=c]{90}{A-sets 2}}} &Random & 0.475 & 0.568 & 0.523 & 0.022 & 0.444 & 0.551 & 0.515 & 0.022 & 0.433 & 0.569 & 0.506 & 0.027 \\ \cline{2-14}\multicolumn{1}{|[3pt]l|}{} & K-Means++ & 0.505 & \textbf{0.598} & 0.543 & 0.02 & 0.483 & 0.584 & 0.547 & 0.024 & 0.473 & 0.58 & 0.531 & 0.021 \\ \cline{2-14}\multicolumn{1}{|[3pt]l|}{} & ROBIN(S) & 0.58 & \textbf{0.598} & 0.591 & 0.008 & 0.581 & \textbf{0.598} & 0.59 & 0.008 & 0.581 & \textbf{0.597} & 0.59 & 0.008 \\ \cline{2-14}\multicolumn{1}{|[3pt]l|}{} & Maximin(S) & 0.504 & 0.574 & 0.54 & 0.019 & 0.5 & 0.58 & 0.536 & 0.022 & 0.49 & 0.575 & 0.536 & 0.022 \\ \cline{2-14}\multicolumn{1}{|[3pt]l|}{} & Kaufman & 0.565 & 0.565 & 0.565 & 0 & 0.565 & 0.565 & 0.565 & 0 & 0.538 & 0.538 & 0.538 & 0 \\ \cline{2-14}\multicolumn{1}{|[3pt]l|}{} & DK-Means++ & 0.598 & \textbf{0.598} & 0.598 & 0 & 0.598 & \textbf{0.598} & 0.598 & 0 & 0.597 & \textbf{0.597} & 0.597 & 0 \\ \cline{2-14}\multicolumn{1}{|[3pt]l|}{} & ROBIN(D) & 0.598 & \textbf{0.598} & 0.598 & 0 & 0.598 & \textbf{0.598} & 0.598 & 0 & 0.597 & \textbf{0.597} & 0.597 & 0 \\ \cline{2-14}\multicolumn{1}{|[3pt]l|}{} & Maximin(D) & 0.555 & 0.555 & 0.555 & 0 & 0.555 & 0.555 & 0.555 & 0 & 0.56 & 0.56 & 0.56 & 0\\ \tabucline [3pt]{-}\multicolumn{1}{|[3pt]l|}{\multirow{8}{*}{\rotatebox[origin=c]{90}{A-sets 3}}} &Random & 0.478 & 0.556 & 0.519 & 0.019 & 0.465 & 0.551 & 0.512 & 0.018 & 0.457 & 0.552 & 0.502 & 0.021 \\ \cline{2-14}\multicolumn{1}{|[3pt]l|}{} & K-Means++ & 0.514 & 0.588 & 0.547 & 0.017 & 0.506 & \textbf{0.601} & 0.548 & 0.017 & 0.485 & 0.576 & 0.533 & 0.019 \\ \cline{2-14}\multicolumn{1}{|[3pt]l|}{} & ROBIN(S) & 0.601 & \textbf{0.601} & 0.601 & 0 & 0.601 & \textbf{0.601} & 0.601 & 0 & 0.601 & \textbf{0.601} & 0.601 & 0 \\ \cline{2-14}\multicolumn{1}{|[3pt]l|}{} & Maximin(S) & 0.525 & 0.585 & 0.556 & 0.016 & 0.525 & 0.589 & 0.558 & 0.015 & 0.52 & 0.586 & 0.559 & 0.017 \\ \cline{2-14}\multicolumn{1}{|[3pt]l|}{} & Kaufman & 0.53 & 0.53 & 0.53 & 0 & 0.53 & 0.53 & 0.53 & 0 & 0.529 & 0.529 & 0.529 & 0 \\ \cline{2-14}\multicolumn{1}{|[3pt]l|}{} & DK-Means++ & 0.601 & \textbf{0.601} & 0.601 & 0 & 0.601 & \textbf{0.601} & 0.601 & 0 & 0.601 & \textbf{0.601} & 0.601 & 0 \\ \cline{2-14}\multicolumn{1}{|[3pt]l|}{} & ROBIN(D) & 0.601 & \textbf{0.601} & 0.601 & 0 & 0.601 & \textbf{0.601} & 0.601 & 0 & 0.601 & \textbf{0.601} & 0.601 & 0 \\ \cline{2-14}\multicolumn{1}{|[3pt]l|}{} & Maximin(D) & 0.588 & 0.588 & 0.588 & 0 & 0.588 & 0.588 & 0.588 & 0 & 0.588 & 0.588 & 0.588 & 0\\ \tabucline [3pt]{-}\multicolumn{1}{|[3pt]l|}{\multirow{8}{*}{\rotatebox[origin=c]{90}{S-sets 1}}} &Random & 0.52 & \textbf{0.711} & 0.616 & 0.037 & 0.545 & 0.663 & 0.612 & 0.035 & 0.497 & 0.662 & 0.587 & 0.047 \\ \cline{2-14}\multicolumn{1}{|[3pt]l|}{} & K-Means++ & 0.58 & \textbf{0.711} & 0.654 & 0.039 & 0.529 & \textbf{0.711} & 0.655 & 0.044 & 0.517 & \textbf{0.711} & 0.651 & 0.044 \\ \cline{2-14}\multicolumn{1}{|[3pt]l|}{} & ROBIN(S) & 0.711 & \textbf{0.711} & 0.711 & 0 & 0.711 & \textbf{0.711} & 0.711 & 0 & 0.711 & \textbf{0.711} & 0.711 & 0 \\ \cline{2-14}\multicolumn{1}{|[3pt]l|}{} & Maximin(S) & 0.611 & \textbf{0.711} & 0.676 & 0.036 & 0.575 & \textbf{0.711} & 0.66 & 0.038 & 0.58 & \textbf{0.711} & 0.648 & 0.038 \\ \cline{2-14}\multicolumn{1}{|[3pt]l|}{} & Kaufman & 0.711 & \textbf{0.711} & 0.711 & 0 & 0.638 & 0.638 & 0.638 & 0 & 0.654 & 0.654 & 0.654 & 0 \\ \cline{2-14}\multicolumn{1}{|[3pt]l|}{} & DK-Means++ & 0.711 & \textbf{0.711} & 0.711 & 0 & 0.711 & \textbf{0.711} & 0.711 & 0 & 0.711 & \textbf{0.711} & 0.711 & 0 \\ \cline{2-14}\multicolumn{1}{|[3pt]l|}{} & ROBIN(D) & 0.711 & \textbf{0.711} & 0.711 & 0 & 0.711 & \textbf{0.711} & 0.711 & 0 & 0.711 & \textbf{0.711} & 0.711 & 0 \\ \cline{2-14}\multicolumn{1}{|[3pt]l|}{} & Maximin(D) & 0.651 & 0.651 & 0.651 & 0 & 0.651 & 0.651 & 0.651 & 0 & 0.652 & 0.652 & 0.652 & 0\\ \tabucline [3pt]{-}\multicolumn{1}{|[3pt]l|}{\multirow{8}{*}{\rotatebox[origin=c]{90}{S-sets 2}}} &Random & 0.464 & \textbf{0.626} & 0.555 & 0.034 & 0.486 & \textbf{0.626} & 0.571 & 0.036 & 0.407 & \textbf{0.626} & 0.53 & 0.055 \\ \cline{2-14}\multicolumn{1}{|[3pt]l|}{} & K-Means++ & 0.516 & \textbf{0.626} & 0.586 & 0.031 & 0.505 & \textbf{0.626} & 0.577 & 0.032 & 0.485 & \textbf{0.626} & 0.566 & 0.035 \\ \cline{2-14}\multicolumn{1}{|[3pt]l|}{} & ROBIN(S) & 0.575 & \textbf{0.626} & 0.617 & 0.02 & 0.575 & \textbf{0.626} & 0.61 & 0.024 & 0.568 & \textbf{0.626} & 0.606 & 0.028 \\ \cline{2-14}\multicolumn{1}{|[3pt]l|}{} & Maximin(S) & 0.533 & \textbf{0.626} & 0.595 & 0.033 & 0.546 & \textbf{0.626} & 0.577 & 0.022 & 0.503 & \textbf{0.626} & 0.569 & 0.027 \\ \cline{2-14}\multicolumn{1}{|[3pt]l|}{} & Kaufman & 0.57 & 0.57 & 0.57 & 0 & 0.57 & 0.57 & 0.57 & 0 & 0.571 & 0.571 & 0.571 & 0 \\ \cline{2-14}\multicolumn{1}{|[3pt]l|}{} & DK-Means++ & 0.626 & \textbf{0.626} & 0.626 & 0 & 0.626 & \textbf{0.626} & 0.626 & 0 & 0.626 & \textbf{0.626} & 0.626 & 0 \\ \cline{2-14}\multicolumn{1}{|[3pt]l|}{} & ROBIN(D) & 0.626 & \textbf{0.626} & 0.626 & 0 & 0.626 & \textbf{0.626} & 0.626 & 0 & 0.626 & \textbf{0.626} & 0.626 & 0 \\ \cline{2-14}\multicolumn{1}{|[3pt]l|}{} & Maximin(D) & 0.529 & 0.529 & 0.529 & 0 & 0.526 & 0.526 & 0.526 & 0 & 0.521 & 0.521 & 0.521 & 0\\ \tabucline [3pt]{-}\multicolumn{1}{|[3pt]l|}{\multirow{8}{*}{\rotatebox[origin=c]{90}{S-sets 3}}} &Random & 0.412 & 0.492 & 0.461 & 0.019 & 0.427 & \textbf{0.493} & 0.463 & 0.018 & 0.395 & \textbf{0.493} & 0.455 & 0.022 \\ \cline{2-14}\multicolumn{1}{|[3pt]l|}{} & K-Means++ & 0.431 & 0.492 & 0.467 & 0.018 & 0.429 & \textbf{0.493} & 0.465 & 0.017 & 0.409 & \textbf{0.493} & 0.459 & 0.02 \\ \cline{2-14}\multicolumn{1}{|[3pt]l|}{} & ROBIN(S) & 0.431 & 0.466 & 0.462 & 0.01 & 0.452 & 0.467 & 0.464 & 0.005 & 0.423 & 0.468 & 0.457 & 0.015 \\ \cline{2-14}\multicolumn{1}{|[3pt]l|}{} & Maximin(S) & 0.431 & 0.492 & 0.469 & 0.018 & 0.427 & 0.492 & 0.469 & 0.019 & 0.422 & \textbf{0.493} & 0.469 & 0.021 \\ \cline{2-14}\multicolumn{1}{|[3pt]l|}{} & Kaufman & 0.492 & 0.492 & 0.492 & 0 & 0.492 & 0.492 & 0.492 & 0 & 0.493 & \textbf{0.493} & 0.493 & 0 \\ \cline{2-14}\multicolumn{1}{|[3pt]l|}{} & DK-Means++ & 0.493 & \textbf{0.493} & 0.493 & 0 & 0.493 & \textbf{0.493} & 0.493 & 0 & 0.493 & \textbf{0.493} & 0.493 & 0 \\ \cline{2-14}\multicolumn{1}{|[3pt]l|}{} & ROBIN(D) & 0.466 & 0.466 & 0.466 & 0 & 0.467 & 0.467 & 0.467 & 0 & 0.464 & 0.464 & 0.464 & 0 \\ \cline{2-14}\multicolumn{1}{|[3pt]l|}{} & Maximin(D) & 0.457 & 0.457 & 0.457 & 0 & 0.464 & 0.464 & 0.464 & 0 & 0.471 & 0.471 & 0.471 & 0\\ \tabucline [3pt]{-}\multicolumn{1}{|[3pt]l|}{\multirow{8}{*}{\rotatebox[origin=c]{90}{S-sets 4}}} &Random & 0.426 & \textbf{0.48} & 0.467 & 0.012 & 0.434 & \textbf{0.48} & 0.465 & 0.012 & 0.4 & \textbf{0.479} & 0.45 & 0.021 \\ \cline{2-14}\multicolumn{1}{|[3pt]l|}{} & K-Means++ & 0.431 & \textbf{0.48} & 0.469 & 0.012 & 0.433 & \textbf{0.48} & 0.469 & 0.012 & 0.412 & \textbf{0.479} & 0.456 & 0.017 \\ \cline{2-14}\multicolumn{1}{|[3pt]l|}{} & ROBIN(S) & 0.457 & \textbf{0.48} & 0.468 & 0.007 & 0.435 & 0.47 & 0.458 & 0.012 & 0.45 & 0.466 & 0.461 & 0.006 \\ \cline{2-14}\multicolumn{1}{|[3pt]l|}{} & Maximin(S) & 0.443 & \textbf{0.48} & 0.47 & 0.007 & 0.456 & 0.471 & 0.468 & 0.004 & 0.438 & \textbf{0.479} & 0.456 & 0.014 \\ \cline{2-14}\multicolumn{1}{|[3pt]l|}{} & Kaufman & 0.48 & \textbf{0.48} & 0.48 & 0 & 0.48 & \textbf{0.48} & 0.48 & 0 & 0.458 & 0.458 & 0.458 & 0 \\ \cline{2-14}\multicolumn{1}{|[3pt]l|}{} & DK-Means++ & 0.48 & \textbf{0.48} & 0.48 & 0 & 0.48 & \textbf{0.48} & 0.48 & 0 & 0.479 & \textbf{0.479} & 0.479 & 0 \\ \cline{2-14}\multicolumn{1}{|[3pt]l|}{} & ROBIN(D) & 0.48 & \textbf{0.48} & 0.48 & 0 & 0.435 & 0.435 & 0.435 & 0 & 0.466 & 0.466 & 0.466 & 0 \\ \cline{2-14}\multicolumn{1}{|[3pt]l|}{} & Maximin(D) & 0.47 & 0.47 & 0.47 & 0 & 0.469 & 0.469 & 0.469 & 0 & 0.462 & 0.462 & 0.462 & 0 \\ \tabucline [3pt]{-}	\end{tabu} } \end{table}

\begin{table}[h]	\centering \caption[ ]{\textbf{Comparison of the initialisation methods on real-world data sets based on Silhouette index.} Each stochastic method (Random, K-Means++, ROBIN(S) and Maximin (S)) was executed 50 times and the minimum, maximum and mean performance is shown followed by the performance variation for three K-Means variations. For the maximum performance the cases where a method has achieved the maximum performance is shown in bold.}  \label{Real} \resizebox{\columnwidth}{!}{	\begin{tabu}{ll| [3pt]c|c|c|c| [3pt]c|c|c|c| [3pt]c|c|c|c| [3pt]} \tabucline [3pt]{3-14} \cline{3-14}& & \multicolumn{4}{c|[3pt]}{K-Means (Hartigan-Wong)} & \multicolumn{4}{c|[3pt]}{K-Means (Lloyd)} & \multicolumn{4}{c|[3pt]}{K-Medians} \\ \cline{3-14}& & min & max & mean & std & min & max & mean & std & min & max & mean & std\\ \tabucline [3pt]{-}\multicolumn{1}{|[3pt]l|}{\multirow{8}{*}{\rotatebox[origin=c]{90}{Iris}}} &Random & 0.517 & \textbf{0.553} & 0.549 & 0.012 & 0.5 & \textbf{0.553} & 0.543 & 0.016 & 0.467 & \textbf{0.551} & 0.542 & 0.016 \\ \cline{2-14}\multicolumn{1}{|[3pt]l|}{} & K-Means++ & 0.517 & \textbf{0.553} & 0.551 & 0.009 & 0.5 & \textbf{0.553} & 0.548 & 0.011 & 0.526 & \textbf{0.551} & 0.55 & 0.006 \\ \cline{2-14}\multicolumn{1}{|[3pt]l|}{} & ROBIN(S) & 0.553 & \textbf{0.553} & 0.553 & 0 & 0.551 & 0.551 & 0.551 & 0 & 0.551 & \textbf{0.551} & 0.551 & 0 \\ \cline{2-14}\multicolumn{1}{|[3pt]l|}{} & Maximin(S) & 0.553 & \textbf{0.553} & 0.553 & 0 & 0.551 & \textbf{0.553} & 0.552 & 0.001 & 0.551 & \textbf{0.551} & 0.551 & 0 \\ \cline{2-14}\multicolumn{1}{|[3pt]l|}{} & Kaufman & 0.553 & \textbf{0.553} & 0.553 & 0 & 0.551 & 0.551 & 0.551 & 0 & 0.551 & \textbf{0.551} & 0.551 & 0 \\ \cline{2-14}\multicolumn{1}{|[3pt]l|}{} & DK-Means++ & 0.553 & \textbf{0.553} & 0.553 & 0 & 0.551 & 0.551 & 0.551 & 0 & 0.551 & \textbf{0.551} & 0.551 & 0 \\ \cline{2-14}\multicolumn{1}{|[3pt]l|}{} & ROBIN(D) & 0.553 & \textbf{0.553} & 0.553 & 0 & 0.551 & 0.551 & 0.551 & 0 & 0.551 & \textbf{0.551} & 0.551 & 0 \\ \cline{2-14}\multicolumn{1}{|[3pt]l|}{} & Maximin(D) & 0.553 & \textbf{0.553} & 0.553 & 0 & 0.553 & \textbf{0.553} & 0.553 & 0 & 0.551 & \textbf{0.551} & 0.551 & 0\\ \tabucline [3pt]{-}\multicolumn{1}{|[3pt]l|}{\multirow{8}{*}{\rotatebox[origin=c]{90}{Ionosphere}}} &Random & 0.296 & \textbf{0.296} & 0.296 & 0 & 0.246 & 0.368 & 0.296 & 0.013 & 0.231 & 0.334 & 0.283 & 0.013 \\ \cline{2-14}\multicolumn{1}{|[3pt]l|}{} & K-Means++ & 0.296 & \textbf{0.296} & 0.296 & 0 & 0.266 & 0.296 & 0.295 & 0.006 & 0.246 & \textbf{0.408} & 0.286 & 0.015 \\ \cline{2-14}\multicolumn{1}{|[3pt]l|}{} & ROBIN(S) & 0.296 & \textbf{0.296} & 0.296 & 0 & 0.296 & 0.296 & 0.296 & 0 & 0.284 & 0.284 & 0.284 & 0 \\ \cline{2-14}\multicolumn{1}{|[3pt]l|}{} & Maximin(S) & 0.296 & \textbf{0.296} & 0.296 & 0 & 0.295 & \textbf{0.408} & 0.314 & 0.038 & 0.284 & \textbf{0.408} & 0.311 & 0.05 \\ \cline{2-14}\multicolumn{1}{|[3pt]l|}{} & Kaufman & 0.296 & \textbf{0.296} & 0.296 & 0 & 0.295 & 0.295 & 0.295 & 0 & 0.284 & 0.284 & 0.284 & 0 \\ \cline{2-14}\multicolumn{1}{|[3pt]l|}{} & DK-Means++ & 0.296 & \textbf{0.296} & 0.296 & 0 & 0.296 & 0.296 & 0.296 & 0 & 0.284 & 0.284 & 0.284 & 0 \\ \cline{2-14}\multicolumn{1}{|[3pt]l|}{} & ROBIN(D) & 0.296 & \textbf{0.296} & 0.296 & 0 & 0.296 & 0.296 & 0.296 & 0 & 0.284 & 0.284 & 0.284 & 0 \\ \cline{2-14}\multicolumn{1}{|[3pt]l|}{} & Maximin(D) & 0.296 & \textbf{0.296} & 0.296 & 0 & 0.296 & 0.296 & 0.296 & 0 & 0.284 & 0.284 & 0.284 & 0\\ \tabucline [3pt]{-}\multicolumn{1}{|[3pt]l|}{\multirow{8}{*}{\rotatebox[origin=c]{90}{Wine}}} &Random & 0.548 & \textbf{0.571} & 0.57 & 0.005 & 0.54 & \textbf{0.571} & 0.57 & 0.005 & 0.566 & \textbf{0.571} & 0.568 & 0.002 \\ \cline{2-14}\multicolumn{1}{|[3pt]l|}{} & K-Means++ & 0.548 & \textbf{0.571} & 0.562 & 0.01 & 0.548 & \textbf{0.571} & 0.566 & 0.007 & 0.566 & \textbf{0.571} & 0.57 & 0.002 \\ \cline{2-14}\multicolumn{1}{|[3pt]l|}{} & ROBIN(S) & 0.571 & \textbf{0.571} & 0.571 & 0 & 0.571 & \textbf{0.571} & 0.571 & 0 & 0.566 & 0.566 & 0.566 & 0 \\ \cline{2-14}\multicolumn{1}{|[3pt]l|}{} & Maximin(S) & 0.553 & \textbf{0.571} & 0.558 & 0.008 & 0.553 & \textbf{0.571} & 0.561 & 0.005 & 0.571 & \textbf{0.571} & 0.571 & 0 \\ \cline{2-14}\multicolumn{1}{|[3pt]l|}{} & Kaufman & 0.571 & \textbf{0.571} & 0.571 & 0 & 0.571 & \textbf{0.571} & 0.571 & 0 & 0.571 & \textbf{0.571} & 0.571 & 0 \\ \cline{2-14}\multicolumn{1}{|[3pt]l|}{} & DK-Means++ & 0.571 & \textbf{0.571} & 0.571 & 0 & 0.571 & \textbf{0.571} & 0.571 & 0 & 0.571 & \textbf{0.571} & 0.571 & 0 \\ \cline{2-14}\multicolumn{1}{|[3pt]l|}{} & ROBIN(D) & 0.571 & \textbf{0.571} & 0.571 & 0 & 0.571 & \textbf{0.571} & 0.571 & 0 & 0.566 & 0.566 & 0.566 & 0 \\ \cline{2-14}\multicolumn{1}{|[3pt]l|}{} & Maximin(D) & 0.548 & 0.548 & 0.548 & 0 & 0.56 & 0.56 & 0.56 & 0 & 0.571 & \textbf{0.571} & 0.571 & 0\\ \tabucline [3pt]{-}\multicolumn{1}{|[3pt]l|}{\multirow{8}{*}{\rotatebox[origin=c]{90}{Breast Cancer}}} &Random & 0.597 & \textbf{0.597} & 0.597 & 0 & 0.597 & \textbf{0.597} & 0.597 & 0 & 0.597 & \textbf{0.597} & 0.597 & 0 \\ \cline{2-14}\multicolumn{1}{|[3pt]l|}{} & K-Means++ & 0.597 & \textbf{0.597} & 0.597 & 0 & 0.597 & \textbf{0.597} & 0.597 & 0 & 0.597 & \textbf{0.597} & 0.597 & 0 \\ \cline{2-14}\multicolumn{1}{|[3pt]l|}{} & ROBIN(S) & 0.597 & \textbf{0.597} & 0.597 & 0 & 0.597 & \textbf{0.597} & 0.597 & 0 & 0.597 & \textbf{0.597} & 0.597 & 0 \\ \cline{2-14}\multicolumn{1}{|[3pt]l|}{} & Maximin(S) & 0.597 & \textbf{0.597} & 0.597 & 0 & 0.597 & \textbf{0.597} & 0.597 & 0 & 0.597 & \textbf{0.597} & 0.597 & 0 \\ \cline{2-14}\multicolumn{1}{|[3pt]l|}{} & Kaufman & 0.597 & \textbf{0.597} & 0.597 & 0 & 0.597 & \textbf{0.597} & 0.597 & 0 & 0.597 & \textbf{0.597} & 0.597 & 0 \\ \cline{2-14}\multicolumn{1}{|[3pt]l|}{} & DK-Means++ & 0.597 & \textbf{0.597} & 0.597 & 0 & 0.597 & \textbf{0.597} & 0.597 & 0 & 0.597 & \textbf{0.597} & 0.597 & 0 \\ \cline{2-14}\multicolumn{1}{|[3pt]l|}{} & ROBIN(D) & 0.597 & \textbf{0.597} & 0.597 & 0 & 0.597 & \textbf{0.597} & 0.597 & 0 & 0.597 & \textbf{0.597} & 0.597 & 0 \\ \cline{2-14}\multicolumn{1}{|[3pt]l|}{} & Maximin(D) & 0.597 & \textbf{0.597} & 0.597 & 0 & 0.597 & \textbf{0.597} & 0.597 & 0 & 0.597 & \textbf{0.597} & 0.597 & 0\\ \tabucline [3pt]{-}\multicolumn{1}{|[3pt]l|}{\multirow{8}{*}{\rotatebox[origin=c]{90}{Glass}}} &Random & 0.192 & 0.456 & 0.43 & 0.056 & 0.194 & 0.452 & 0.345 & 0.09 & 0.137 & 0.442 & 0.268 & 0.077 \\ \cline{2-14}\multicolumn{1}{|[3pt]l|}{} & K-Means++ & 0.271 & \textbf{0.587} & 0.457 & 0.053 & 0.278 & \textbf{0.585} & 0.454 & 0.064 & 0.211 & \textbf{0.592} & 0.432 & 0.089 \\ \cline{2-14}\multicolumn{1}{|[3pt]l|}{} & ROBIN(S) & 0.447 & 0.452 & 0.447 & 0.001 & 0.43 & 0.448 & 0.443 & 0.005 & 0.257 & 0.397 & 0.384 & 0.028 \\ \cline{2-14}\multicolumn{1}{|[3pt]l|}{} & Maximin(S) & 0.555 & \textbf{0.587} & 0.582 & 0.009 & 0.43 & \textbf{0.585} & 0.563 & 0.048 & 0.433 & \textbf{0.592} & 0.576 & 0.03 \\ \cline{2-14}\multicolumn{1}{|[3pt]l|}{} & Kaufman & 0.452 & 0.452 & 0.452 & 0 & 0.448 & 0.448 & 0.448 & 0 & 0.238 & 0.238 & 0.238 & 0 \\ \cline{2-14}\multicolumn{1}{|[3pt]l|}{} & DK-Means++ & 0.447 & 0.447 & 0.447 & 0 & 0.431 & 0.431 & 0.431 & 0 & 0.435 & 0.435 & 0.435 & 0 \\ \cline{2-14}\multicolumn{1}{|[3pt]l|}{} & ROBIN(D) & 0.447 & 0.447 & 0.447 & 0 & 0.444 & 0.444 & 0.444 & 0 & 0.392 & 0.392 & 0.392 & 0 \\ \cline{2-14}\multicolumn{1}{|[3pt]l|}{} & Maximin(D) & 0.584 & 0.584 & 0.584 & 0 & 0.583 & 0.583 & 0.583 & 0 & 0.58 & 0.58 & 0.58 & 0\\ \tabucline [3pt]{-}\multicolumn{1}{|[3pt]l|}{\multirow{8}{*}{\rotatebox[origin=c]{90}{Yeast}}} &Random & 0.142 & 0.155 & 0.155 & 0.011 & 0.136 & 0.156 & 0.155 & 0.011 & 0.117 & 0.175 & 0.144 & 0.012 \\ \cline{2-14}\multicolumn{1}{|[3pt]l|}{} & K-Means++ & 0.15 & 0.217 & 0.177 & 0.012 & 0.148 & 0.192 & 0.173 & 0.01 & 0.119 & 0.184 & 0.157 & 0.013 \\ \cline{2-14}\multicolumn{1}{|[3pt]l|}{} & ROBIN(S) & 0.18 & 0.183 & 0.181 & 0.001 & 0.178 & 0.19 & 0.183 & 0.006 & 0.161 & 0.172 & 0.164 & 0.005 \\ \cline{2-14}\multicolumn{1}{|[3pt]l|}{} & Maximin(S) & 0.189 & \textbf{0.224} & 0.202 & 0.01 & 0.173 & \textbf{0.225} & 0.204 & 0.014 & 0.166 & \textbf{0.213} & 0.194 & 0.018 \\ \cline{2-14}\multicolumn{1}{|[3pt]l|}{} & Kaufman & 0.161 & 0.161 & 0.161 & 0 & 0.159 & 0.159 & 0.159 & 0 & 0.151 & 0.151 & 0.151 & 0 \\ \cline{2-14}\multicolumn{1}{|[3pt]l|}{} & DK-Means++ & 0.155 & 0.155 & 0.155 & 0 & 0.156 & 0.156 & 0.156 & 0 & 0.14 & 0.14 & 0.14 & 0 \\ \cline{2-14}\multicolumn{1}{|[3pt]l|}{} & ROBIN(D) & 0.183 & 0.183 & 0.183 & 0 & 0.19 & 0.19 & 0.19 & 0 & 0.172 & 0.172 & 0.172 & 0 \\ \cline{2-14}\multicolumn{1}{|[3pt]l|}{} & Maximin(D) & 0.192 & 0.192 & 0.192 & 0 & 0.191 & 0.191 & 0.191 & 0 & 0.175 & 0.175 & 0.175 & 0 \\ \tabucline [3pt]{-}	\end{tabu} } \end{table}

\cleardoublepage

\begin{figure}
	\centering
	\includegraphics[width=\linewidth]{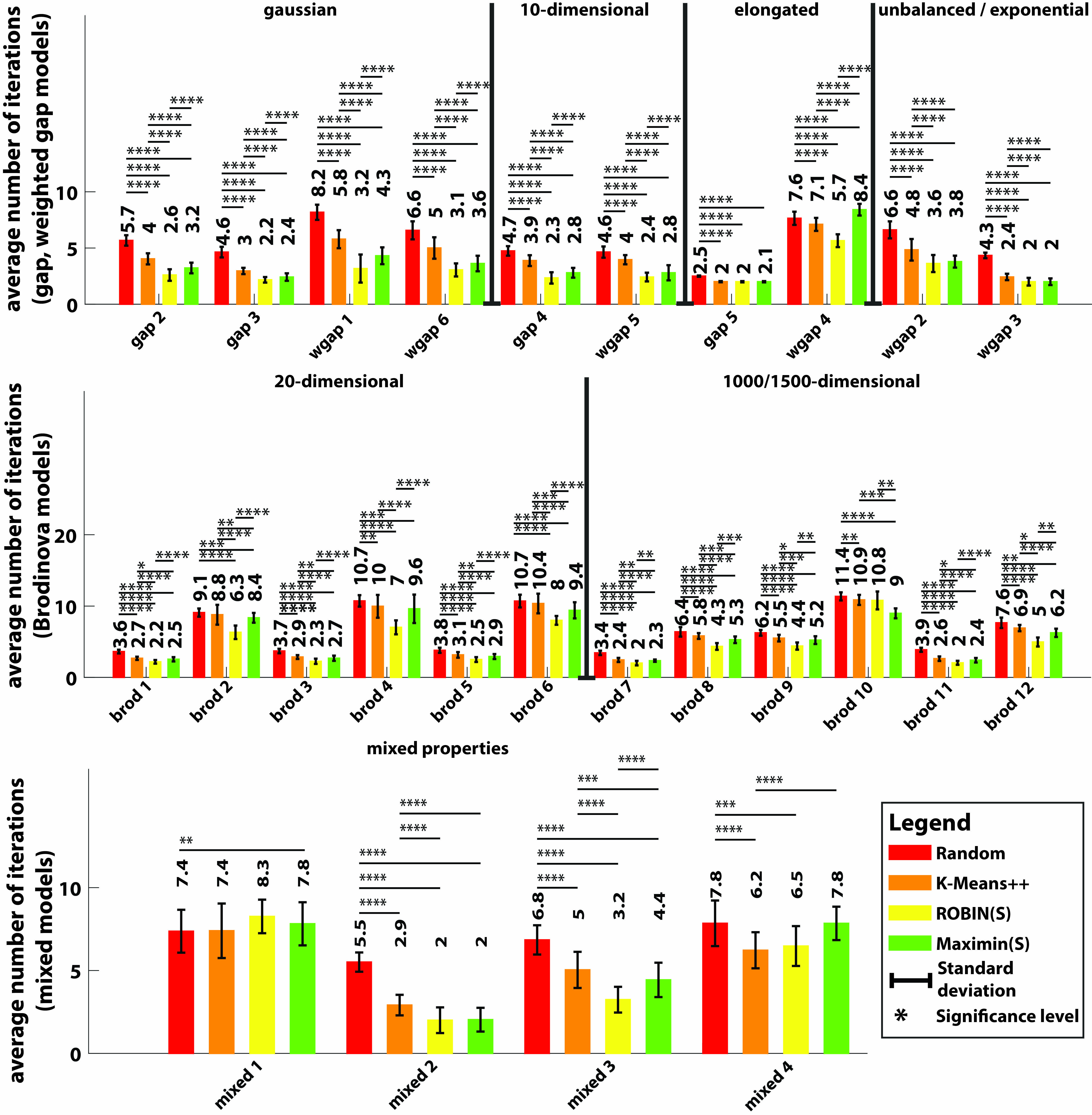}
	\caption{\textbf{Average number of iterations until convergence for the stochastic methods.} Each plot shows the number of iteration of the Lloyd's K-Means algorithm (y-axis) until it reaches convergence using different stochastic initialisation methods on different data sets models (x-axis). To calculate the average number of iterations, we averaged the number of iterations across the 25 runs on the 40 data sets for each model (gap, weighted gap, Brodinova and mixed). The standard deviation corresponds to the average standard deviation over the 25 runs of each data set. Solid lines on any two bars underline the level of significant difference between the corresponding methods (cases of no significant differences are not showing). Table \ref{allps} shows a summary of the comparisons among all the different initialisation methods.} \label{figS}  
\end{figure}

\cleardoublepage

\begin{figure}
	\centering
	\includegraphics[width=\linewidth]{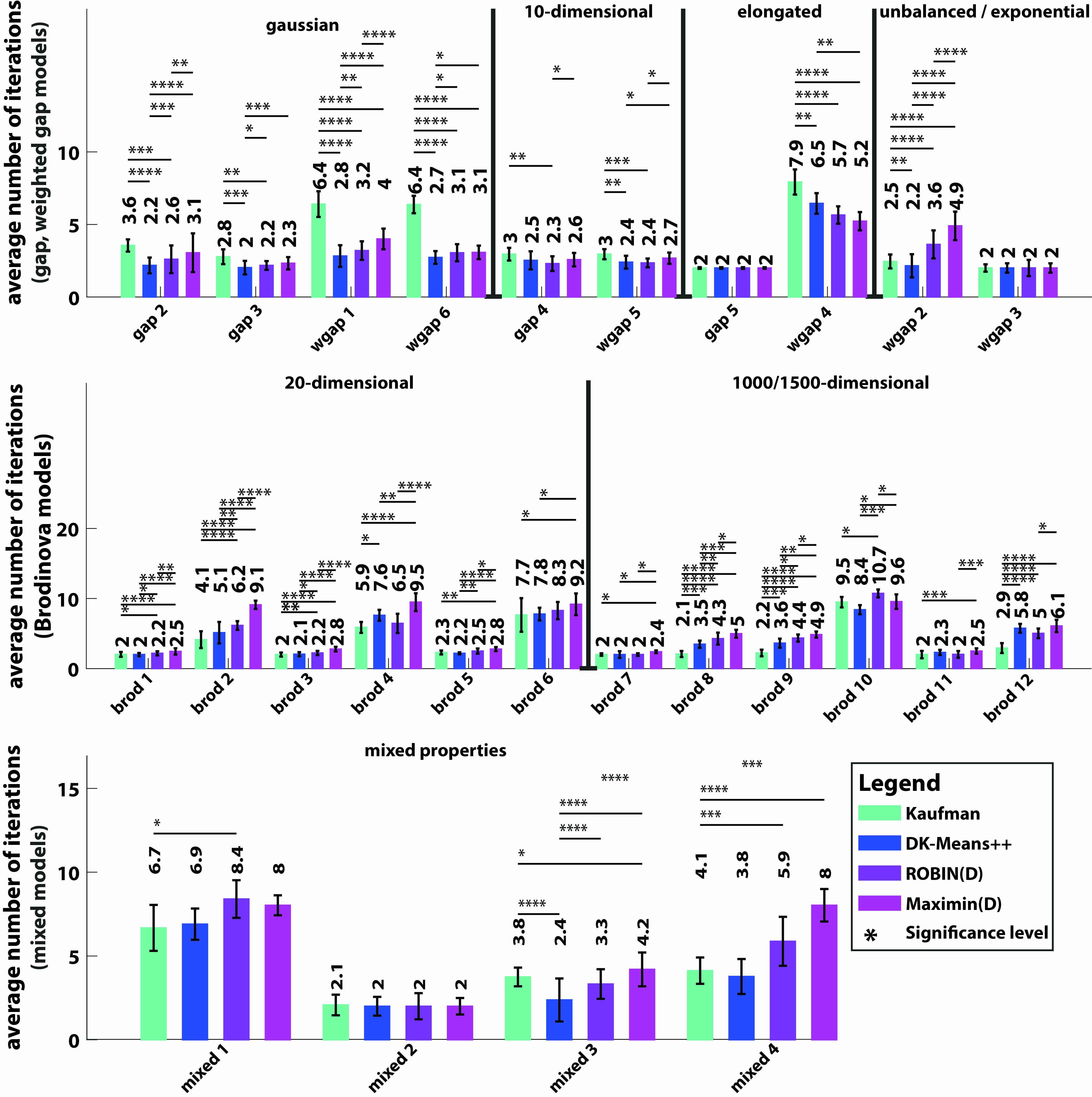}
	\caption{\textbf{Average number of iterations until convergence for the stochastic methods.} Each plot shows the number of iteration of the Lloyd's K-Means algorithm (y-axis) until it reaches convergence using different deterministic initialisation methods on different data sets models (x-axis). To calculate the average number of iterations, we averaged the number of iterations across the 25 runs on the 40 data sets for each model (gap, weighted gap, Brodinova and mixed). The standard deviation corresponds to the average standard deviation over the 25 runs of each data set. Solid lines on any two bars underline the level of significant difference between the corresponding methods (cases of no significant differences are not showing). Table \ref{allps} shows a summary of the comparisons among all the different initialisation methods.} \label{figD}  
\end{figure}

\cleardoublepage

\begin{table}
	\centering
	\begin{tabular}{|l|c|c|}
		\hline
		\multicolumn{1}{|c|}{\begin{tabular}[c]{@{}c@{}}Initialization\\ method\end{tabular}} & \begin{tabular}[c]{@{}c@{}}Total number\\ of instances\end{tabular} & \begin{tabular}[c]{@{}c@{}}Significantly more iterations\\ for K-Means to converge\end{tabular} \\ \hline
		Random vs K-Means++                                                                   & 26                                                                  & 23 vs 0                                                                            \\ \hline
		Random vs ROBIN(S)                                                                    & 26                                                                  & 24 vs 0                                                                            \\ \hline
		Random vs Kaufman                                                                     & 26                                                                  & 23 vs 0                                                                            \\ \hline
		Random vs DK-Means++                                                                  & 26                                                                  & 25 vs 0                                                                            \\ \hline
		Random vs ROBIN(D)                                                                    & 26                                                                  & 24 vs 0                                                                            \\ \hline
		Random vs Maximin(S)                                                                  & 26                                                                  & 23 vs 2                                                                            \\ \hline
		Random vs Maximin(D)                                                                  & 26                                                                  & 23 vs 0                                                                            \\ \hline
		K-Means++ vs ROBIN(S)                                                                 & 26                                                                  & 22 vs 0                                                                            \\ \hline
		K-Means++ vs Kaufman                                                                  & 26                                                                  & 21 vs 2                                                                            \\ \hline
		K-Means++ vs DK-Means++                                                               & 26                                                                  & 24 vs 0                                                                            \\ \hline
		K-Means++ vs ROBIN(D)                                                                 & 26                                                                  & 22 vs 0                                                                            \\ \hline
		K-Means++ vs Maximin(S)                                                               & 26                                                                  & 21 vs 3                                                                            \\ \hline
		K-Means++ vs Maximin(D)                                                               & 26                                                                  & 19 vs 1                                                                            \\ \hline
		ROBIN(S) vs Kaufman                                                                   & 26                                                                  & 11 vs 7                                                                            \\ \hline
		ROBIN(S) vs DK-Means++                                                                & 26                                                                  & 14 vs 0                                                                            \\ \hline
		ROBIN(S) vs ROBIN(D)                                                                  & 26                                                                  & 3 vs 2                                                                             \\ \hline
		ROBIN(S) vs Maximin(S)                                                                & 26                                                                  & 1 vs 21                                                                            \\ \hline
		ROBIN(S) vs Maximin(D)                                                                & 26                                                                  & 1 vs 19                                                                            \\ \hline
		Kaufman vs DK-Means++                                                                 & 26                                                                  & 8 vs 4                                                                             \\ \hline
		Kaufman vs ROBIN(D)                                                                   & 26                                                                  & 7 vs 10                                                                            \\ \hline
		Kaufman vs Maximin(S)                                                                 & 26                                                                  & 3 vs 15                                                                            \\ \hline
		Kaufman vs Maximin(D)                                                                 & 26                                                                  & 3 vs 14                                                                            \\ \hline
		DK-Means++ vs ROBIN(D)                                                                & 26                                                                  & 0 vs 15                                                                            \\ \hline
		DK-Means++ vs Maximin(S)                                                              & 26                                                                  & 0 vs 20                                                                            \\ \hline
		DK-Means++ vs Maximin(D)                                                              & 26                                                                  & 1 vs 18                                                                            \\ \hline
		ROBIN(D) vs Maximin(S)                                                                & 26                                                                  & 1 vs 21                                                                            \\ \hline
		ROBIN(D) vs Maximin(D)                                                                & 26                                                                  & 1 vs 17                                                                            \\ \hline
		Maximin(S) vs Maximin(D)                                                              & 26                                                                  & 8 vs 1                                                                             \\ \hline
	\end{tabular}
	\caption{\textbf{Summary of comparisons for the number of iterations until convergence for the Lloyd's K-Means algorithm using different initialisation methods.} Each row shows a comparison between different initialisation methods on the number of times that each method resulted on the Lloyd's K-Means algorithm to have greater number of iterations until convergence. As indicator of performance, a method resulted to lower number of iterations is consider better, thus the lower the score the better the methods. Based on the results, ROBIN(S) is the best stochastic method and results to the less iterations for the K-Means algorithm to reach convergence; DK-Means++ is the best deterministic method and results to the less iterations for the K-Means algorithm to reach convergence. Overall, deterministic methods result to lower number of iterations for the K-Means algorithm.} \label{allps}  
\end{table}

\cleardoublepage

%\cleardoublepage

\begin{figure}
	\centering
	\includegraphics[width=\linewidth]{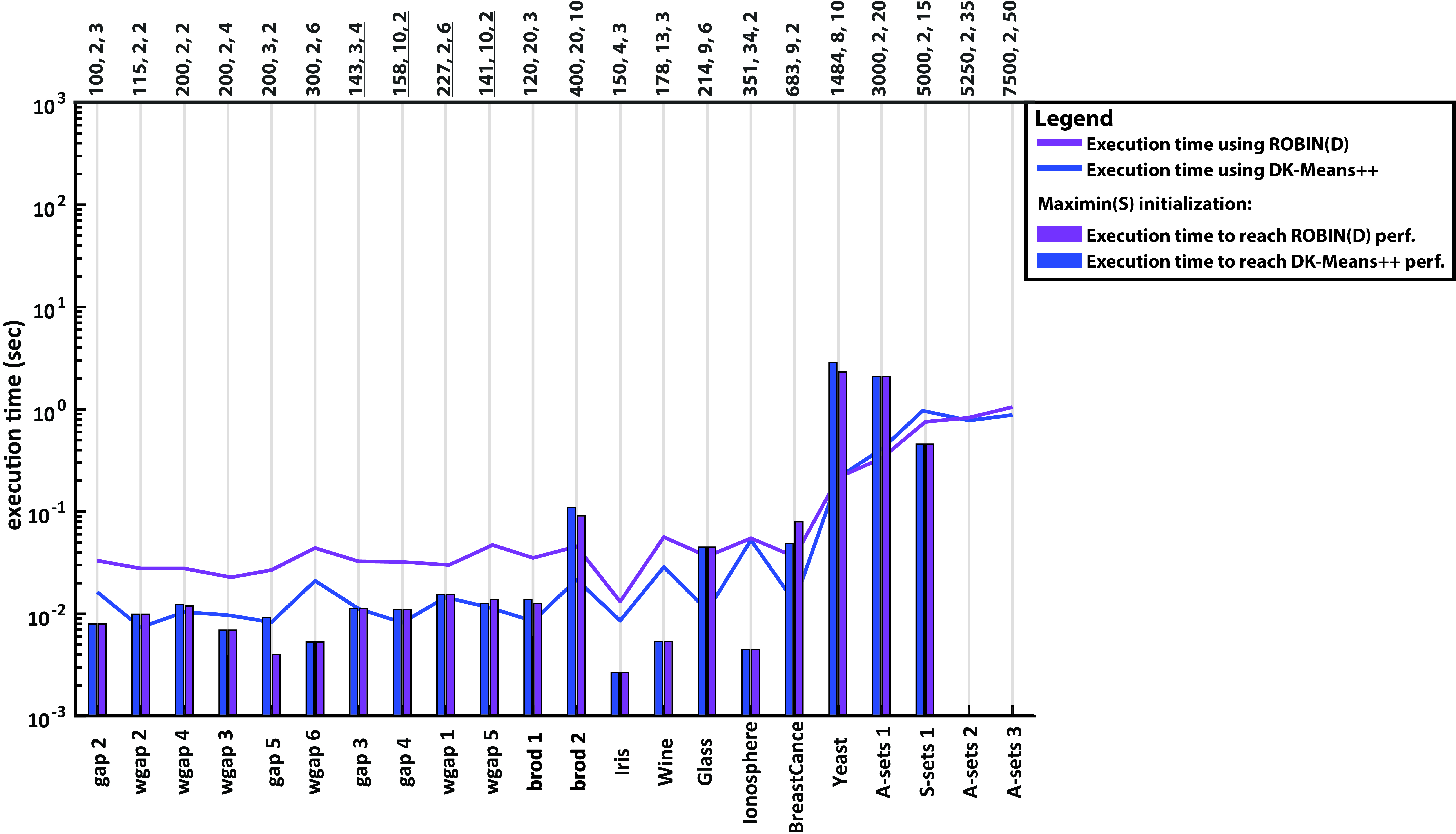}
	\caption{\textbf{Execution time analysis for K-Means clustering with Maximin(S) initialisation to reach the performance of DK-Means++ and ROBIN(D).} Each bar shows the execution duration of K-Means clustering algorithm executed multiple times with the Maximin(S) initialisation method until reaching the performance of the clustering algorithm running once using DK-Means++ and ROBIN(D) methods. Lines indicate the execution time of K-Means clustering running once using the deterministic methods ROBIN(D) and DK-Means++. The clustering algorithm was given 50 execution repetitions with the stochastic Maximin(S) method to reach an equal or better solution than deterministic methods. Data sets without a bar means that no equal or better solution compared to the respective clustering with a deterministic method was found. The data sets are arranged based on their \textit{size}, \textit{dimensionality} and \textit{number of clusters} (see info on top, underlined numbers means that for these models the generated data sets had different sizes). Results were averaged over 40 data sets for the data set models.} \label{timeMaximin} 
\end{figure}

\cleardoublepage

\section*{Details about the Gaussian data sets of gap and weighted gap studies}

\begin{itemize}
	\item gap model 2: means of the cluster centers are as follows: [0,0], [0,5], [5,-3]. Standard deviation of all the centers is 1.
	\item gap model 3: means of the cluster centers are all randomly selected as N(0, 5$I$).
	\item weighted gap model 1: means of the cluster centers are as follows: [10,0], [6,0], [0,0], [−5, 0], [5, 5], [0, −6]. Standard deviation of all the centers is 1.
	\item weighted gap model 6: means of the cluster centers are as follows: [0,0], [-1,5], [10,-10], [15,-10], [10,-15], [25,25]. Standard deviation of all the centers is 1.
	\item gap model 4: means of the cluster centers are all randomly selected as N(0, 1.9$I$).
	\item weighted gap model 5: means of the cluster centers are all randomly selected as N(0, 3.6$I$).	
	\item gap model 5: clusters where consisted of 100 points generated from 3 features with values equally spaced from -0.5 to 0.5 and the addition of Gaussian noise with standard deviation of 0.1. The values on all three dimensions of the second cluster was then increased by 10. 
	\item weighted gap model 4: Similar to gap model 5 but the values of the second cluster were decreased by 1.
\end{itemize}	

\noindent $I$ refers to the identity matrix with size equal to the dimensionality of the data set. Any data set with distance between clusters less than 1 was discarded and regenerated.

\end{document}